\documentclass[lettersize,journal]{IEEEtran}
\usepackage{algorithmic}
\usepackage{array}
\usepackage[caption=false,font=normalsize,labelfont=sf,textfont=sf]{subfig}
\usepackage{textcomp}
\usepackage{stfloats}
\usepackage{url}
\usepackage{verbatim}
\usepackage{graphicx}
\usepackage{rotating}  

\usepackage{multirow,bbding}
\usepackage{amsmath,amssymb,amsfonts,bm}
\usepackage{threeparttable}

\usepackage{booktabs}

\usepackage{tcolorbox}
\usepackage{mdframed}

\usepackage{wasysym}

\usepackage{cite}

\usepackage{tablefootnote}

\usepackage{longtable}

\usepackage{tikz}
\usepackage{xcolor}

\usetikzlibrary{shapes,arrows,positioning,fit,backgrounds}

\usepackage{hyperref}
\hypersetup{
hidelinks
}

\usepackage{fontawesome5}

\definecolor{lime}{HTML}{A6CE39}
\DeclareRobustCommand{\orcidicon}{%
\begin{tikzpicture}
\draw[lime, fill=lime] (0,0) 
circle [radius=0.16] 
node[white] {{\fontfamily{qag}\selectfont \tiny ID}};\draw[white, fill=white] (-0.0625,0.095) 
circle [radius=0.007];\end{tikzpicture}
\hspace{-2mm}}
\foreach \x in {A, ..., Z}{%
\expandafter\xdef\csname orcid\x\endcsname{\noexpand\href{https://orcid.org/\csname orcidauthor\x\endcsname}{\noexpand\orcidicon}}
}

\def\BibTeX{{\rm B\kern-.05em{\sc i\kern-.025em b}\kern-.08em
T\kern-.1667em\lower.7ex\hbox{E}\kern-.125emX}}
\usepackage{balance}
\begin{document}

\tikzstyle{box} = [rectangle, minimum width=3cm, minimum height=1cm, text centered, draw=black]
\tikzstyle{arrow} = [thick,->,>=stealth]

\usetikzlibrary{positioning}

\title{Privacy-Preserving Video Anomaly Detection: A Survey}

\author{
\centering
Yang Liu\orcidA{},
Siao Liu\orcidK{}, 
Xiaoguang Zhu\orcidB{}, 
Jielin Li\orcidG{}, 
Hao Yang\orcidC{}, 
Liangyu Teng\orcidD{}, 
Junchen Guo\orcidE{}, 
Yan Wang\orcidH{},  
Dingkang Yang\orcidL{}, 
Jing Liu\orcidF{}

\thanks{
  This work was supported in part by the National Natural Science Foundation of China under Grant 62406075, National Key Research and Development Program of China under Grant 2023YFC3604802, and the China Postdoctoral Science Foundation under Grant 2023M730647 and Grant 2023TQ0075. This work was also supported by Mitacs Elevate under Grant IT44479, Canada. \textit{(Corresponding authors: Yan Wang, Dingkang Yang, and Jing Liu.)}
}

\thanks{Yang Liu is with the Department of Computer Science, University of Toronto, Ontario, M5S 1A1, Canada (email: yangliu@cs.toronto.edu).}

\thanks{Siao Liu, Hao Yang, Liangyu Teng, Juncen Guo, and Dingkang Yang are with the College of Intelligent Robotics and Advanced Manufacturing,  Fudan University, Shanghai, 200433, China (emails: saliu20@fudan.edu.cn, yanghao21@m.fudan.edu.cn, lyteng20@fudan.edu.cn, guojc23@m.fudan.edu.cn, dkyang20@fudan.edu.cn).}

\thanks{Xiaoguang Zhu is with the DataLab: Data Science and Informatics, University of California, Davis, California, 95616, USA (email: xgzhu@ucdavis.edu).}

\thanks{Jielin Li is with the Department of Computer Science, The University of Hong Kong, Hong Kong, 999077, China (email: jielinli@connect.hku.hk).}

\thanks{Yan Wang is with the School of Data Science and Engineering, East China Normal University, Shanghai, 200062, China (email: yanwang@dase.ecnu.edu.cn).}

\thanks{Jing Liu is with the College of Future Information Technology, Fudan University, Shanghai 200433, China, also with the Division of Natural and Applied Sciences, Duke Kunshan University, Suzhou 215316, China, and also with the Department of Electrical and Computer Engineering, The University of British Columbia, BC, V6T 1Z4, Canada (e-mail: jing.liu@ieee.org).}

}

\markboth{IEEE Transactions on Neural Networks and Learning Systems,~Vol.~xx, No.~xx, xxxx~xxxx}%
{Privacy-Preserving Video Anomaly Detection: A Survey}

\maketitle

\begin{abstract}

Video Anomaly Detection (VAD) aims to automatically analyze spatiotemporal patterns in surveillance videos collected from open spaces to detect anomalous events that may cause harm, such as fighting, stealing, and car accidents. However, vision-based surveillance systems such as closed-circuit television often capture personally identifiable information. The lack of transparency and interpretability in video transmission and usage raises public concerns about privacy and ethics, limiting the real-world application of VAD. Recently, researchers have focused on privacy concerns in VAD by conducting systematic studies from various perspectives including data, features, and systems, making Privacy-Preserving Video Anomaly Detection (P2VAD) a hotspot in the AI community. However, current research in P2VAD is fragmented, and prior reviews have mostly focused on methods using RGB sequences, overlooking privacy leakage and appearance bias considerations. To address this gap, this article is the first to systematically reviews the progress of P2VAD, defining its scope and providing an intuitive taxonomy. We outline the basic assumptions, learning frameworks, and optimization objectives of various approaches, analyzing their strengths, weaknesses, and potential correlations. Additionally, we provide open access to research resources such as benchmark datasets and available code. Finally, we discuss key challenges and future opportunities from the perspectives of AI development and P2VAD deployment, aiming to guide future work in the field.

\end{abstract}


\section{Introduction}~\label{sec1}

\IEEEPARstart{V}{ideo} Anomaly Detection (VAD) aims to automatically identify irregular patterns in spatio-temporal surveillance data to detect unexpected anomalous events \cite{liu2024networking}. Due to its ability to capture real-time environmental information in open spaces, VAD has demonstrated promising applications in emerging fields such as smart cities \cite{huang2022self}, modern industry \cite{amp}, and healthcare \cite{galvao2024anomaly}. Applications include traffic accident warning in Intelligent Transportation Systems (ITS) \cite{liu2023distributional}, identifying irregularities in industrial production \cite{liu2024memory}, and detecting elderly falls. Compared to Action Recognition (AR) \cite{10884961,sun2022human}, which relies on fine-grained labels for model training, VAD generally follows the open-world assumption that real-world anomalies are rare and unbounded \cite{pang2021deep}. As a result, collecting all possible anomalous events in diverse and dynamic scenarios for training supervised multi-class classification models is impractical. Moreover, AR methods are ill-suited to handle the significant data imbalance and label noise often encountered in anomaly detection. In contrast, VAD is often formulated as an unsupervised outlier detection task, using easily collected regular events to train the model to characterize the prototypical patterns of normal videos, while treating uncharacterizable samples as anomalies \cite{zhang2020normality}. Thus, Unsupervised VAD (UVAD) avoids the expensive data preparation cost of collecting anomalies and exhibits great scene adaptability, with the theoretical ability to detect any positive samples distinct from negative ones \cite{wang2025dual,cheng2023learning,liu2023stochastic,zhao2022exploiting,li2024stnmamba}.

In recent years, researchers have proposed Weakly-supervised VAD (WsVAD) approaches, which utilize video-level labeled regression models to compute fine-grained frame-level anomaly scores for the temporal localization of anomalous events \cite{zhou2024batchnorm,wang2025federated}. Although such approaches \cite{MIR,MIST,RTFM} are usually limited to identifying predefined anomaly categories in a given scene, they still do not require segment-level or frame-level labeling, but instead use binary labels (0 or 1) to indicate whether a video contains anomalies or not, without the need for second-level categorization. Existing work \cite{MIST} has shown that weakly-supervised VAD models do not need to consider the data balance between various types of anomalous samples and regular samples when identifying anomalous behaviors that may threaten life safety or cause economic loss in specific scenarios (e.g., criminal events and violent behaviors). These models have proven to be more reliable and practical than weakly-supervised action recognition models that rely on multi-class labels. The latest VAD research aims to detect anomalies directly from raw, unfiltered surveillance data. The so-called Fully-unsupervised VAD (FuVAD) allows models to be automatically trained using data from large-scale video IoT systems and online video platforms.

However, existing VAD studies \cite{huang2022weakly,ning2024memory,PC-LSTM,LSTM-AE,STU-net,GANs,osin} typically use RGB video sequences that contain identifiable and sensitive environmental information as inputs for modeling, raising public concerns about individual privacy, appearance bias, and data security. These considerations are particularly prominent in sectors such as healthcare and the military, where privacy issues not only limit the adoption of VAD technology but also erode public trust and impede research progress. The collection, storage, and use of RGB sequences containing facial data, clothing information, and specific environmental layouts, especially through Closed-Circuit Tele-Vision (CCTV) or online internet platforms, can be offensive, and even provoke fear of personal information misuse. Furthermore, models typically use deep neural networks that directly operate on original high-definition color sequences for normality learning and anomaly detection, making it difficult for the public to trust VAD technology due to the black-box nature of data processing, which lacks interpretability and transparency \cite{yang2024context,cao2024context,tan2024overlooked}.

More importantly, VAD systems deployed in real-world applications often involve tens of thousands of clients across large-scale spaces such as neighborhoods, buildings, or entire cities. While the large-scale, 24/7 video streams generated across diverse scenarios are beneficial for training data-driven deep models, the data exchanges between sensor endpoints, edge computing units, and data centers make privacy and security critical concerns, often overshadowing detection accuracy. Unfortunately, despite the great social value and application potential of Privacy-Preserving VAD (P2VAD), the absence of a clear development lineage and standardized validation benchmarks in existing research has led to fragmented efforts, as researchers from different backgrounds tend to focus on different issues, which limits the continuous progress and real-world applications of this technology.

In this regard, this article reviews existing VAD work from a privacy-preserving perspective and provides the first P2VAD taxonomy. We highlight the core concerns and fundamental ideas of different research directions, as well as compile all publicly available benchmark datasets and evaluation metrics, with the aim of standardizing future research and promoting the development of trusted P2VAD applications.

\subsection{Related Surveys}

Although several review papers \cite{nayak2021comprehensive,rezaee2021survey,santhosh2020anomaly,ramachandra2020survey,chandrakala2022anomaly,liu2024generalized} on VAD have been published over the past three years, they primarily focus on categorizing methods based on RGB videos. While these works propose various insightful VAD classification systems for different scenarios and orientations, they do not sufficiently address privacy protection and system security in VAD research. Table~\ref{t-rela} summarizes the main focus areas, methodologies, and perspectives of recent reviews related to anomaly detection. Many of them \cite{nayak2021comprehensive,rezaee2021survey,santhosh2020anomaly} approach VAD from a narrow perspective, often considering it as a sub-task within the broader fields of anomaly detection or video understanding, thereby neglecting its interdisciplinary aspects. Even recent surveys \cite{chandrakala2022anomaly,liu2024generalized,liu2024networking} do not emphasize privacy concerns, instead limiting their scope to algorithm categorization.

\begin{table}[]
  \centering
  \caption{Comparison with related surveys.}
  \label{t-rela}
  \begin{threeparttable}
  \resizebox{\linewidth}{!}{
  \begin{tabular}{@{}llccccccc@{}}
  \toprule
  \multirow{2}{*}{\textbf{Venue}} & \multicolumn{1}{c}{\multirow{2}{*}{\textbf{Main Focus}}} & \multirow{2}{*}{\textbf{P2}} & \multicolumn{3}{c}{\textbf{Pathways Discussed}} & \multicolumn{3}{c}{\textbf{Viewpoint}} \\ \cmidrule(l){4-6} \cmidrule(l){7-9}  
  & \multicolumn{1}{c}{}   & & UVAD   & WsVAD  & FuVAD & \textit{AD}  & \textit{VU}  & \textit{C}  \\ \midrule
  IVC21 \cite{nayak2021comprehensive}   & Deep unsupervised VAD  & \XSolidBrush &  \CIRCLE  & \Circle & \Circle& \smiley & \frownie & \frownie\\
  PUC21 \cite{rezaee2021survey}  & Real-time crowd VAD& \XSolidBrush &  \CIRCLE  & \Circle & \Circle& \smiley & \frownie & \frownie\\
  CSUR21 \cite{santhosh2020anomaly}  & VAD in traffic scene   & \XSolidBrush &  \LEFTcircle  &  \LEFTcircle  & \Circle& \frownie & \smiley & \frownie\\
  CSUR22 \cite{pang2021deep}  & Deep anomaly detection & \XSolidBrush & \astrosun  & \astrosun  & \astrosun & \smiley & \frownie & \frownie\\
  TPAMI22 \cite{ramachandra2020survey} & Single scene VAD   & \XSolidBrush &  \CIRCLE &  \LEFTcircle  & \Circle& \smiley & \smiley & \frownie \\
  AIR23 \cite{chandrakala2022anomaly}  & One- and two-class VAD & \XSolidBrush &  \CIRCLE &  \LEFTcircle  & \Circle& \smiley & \smiley & \frownie \\
  CSUR23 \cite{liu2024generalized} & Generalized VAD& \XSolidBrush &  \CIRCLE &  \CIRCLE &  \CIRCLE& \smiley & \smiley & \frownie\\
  TETCL24 \cite{mishra2024skeletal} & Deep skeleton VAD  & \XSolidBrush &  \CIRCLE &  \LEFTcircle  & \Circle & \frownie & \smiley & \frownie\\
  CSUR25 \cite{liu2024networking} & Networking systems for VAD  & \XSolidBrush &  \CIRCLE &  \CIRCLE &  \CIRCLE& \smiley & \smiley & \smiley    \\
  \midrule
  Ours& Privacy-Preserving VAD & \Checkmark &  \CIRCLE &  \CIRCLE &  \CIRCLE& \smiley & \smiley & \smiley    \\ \bottomrule
  \end{tabular}
  }
  \begin{tablenotes}
  \footnotesize
  \item \CIRCLE : Systematic presentation. \LEFTcircle : Briefly mentioned. \Circle : Not presented. 
  \item \astrosun : Not applicable. \smiley/\frownie : Research perspective involved/unincluded.
  \item \textit{AD} = Anomaly Detection, \textit{VU} = Video Understanding, \textit{C} = Computing.
    \end{tablenotes} 
  \end{threeparttable}
\vspace{-10pt}
\end{table}

To be more specific, early reviews either focused on specific applications such as transportation \cite{santhosh2020anomaly} or categorized models based on different learning paradigms \cite{chandrakala2022anomaly,liu2024generalized}, without addressing data security issues. For example, Liu et al. \cite{liu2024generalized} discussed unsupervised, weakly supervised, and fully supervised approaches, classifying P2VAD under these categories based on supervision signals, while overlooking its core contribution to privacy protection. Although some works mention the importance of data security for future VAD development and suggest that trustworthy systems will become a mainstream trend, they offer limited insights into P2VAD research due to a lack of detailed explanations and targeted literature.

In 2024, Mishra et al. \cite{mishra2024skeletal} introduced the first review that focused on skeleton-based VAD, highlighting the advantage of gesture keypoints over RGB video in privacy preservation, indicating that P2VAD is becoming a significant research area. However, their focus was limited to a specific sub-task—modeling video normality using skeleton data—without integrating broader privacy-preserving efforts across data acquisition, model learning, and system-level applications. A comprehensive review that addresses these aspects together would provide a more holistic perspective and standardize future P2VAD research.

\subsection{P2VAD Taxonomy}

Privacy-preserving video anomaly detection emerges from the recognition that conventional VAD systems, while effective in detecting anomalous behaviors, inherently pose significant privacy risks by processing RGB video sequences that contain identifiable human appearance information, facial features, clothing details, and environmental layouts. These privacy concerns have become increasingly critical as surveillance systems expand across public and private spaces, necessitating the development of VAD approaches that can maintain detection effectiveness while protecting individual privacy and complying with data protection regulations. 

\begin{figure}[t]
  \centering
  \includegraphics[width=\linewidth]{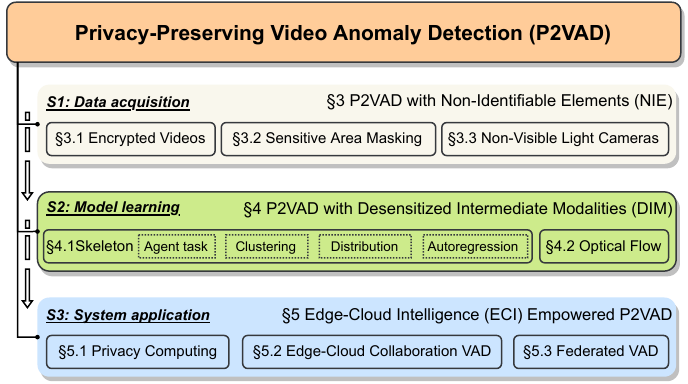}
  \caption{Taxonomy of Privacy-Preserving Video Anomaly Detection (P2VAD).}
  \label{f-taxo}
\vspace{-10pt}
\end{figure}

To address these challenges systematically, this article introduces the first comprehensive VAD taxonomy from a privacy-preserving perspective, as illustrated in Fig.~\ref{f-taxo}. The taxonomy is constructed around the fundamental understanding that privacy risks in VAD systems manifest at different stages of the data lifecycle, requiring distinct mitigation strategies. Privacy breaches can occur during initial data acquisition when cameras capture identifiable information, during model learning when algorithms process sensitive appearance features, and during system deployment when data is transmitted across distributed computing infrastructures.

The proposed P2VAD taxonomy encompasses all related works across three critical stages corresponding to the data lifecycle: \textit{S1: data acquisition}; \textit{S2: model learning}; and \textit{S3: system applications}. These stages naturally lead to three main categories of privacy-preserving approaches:
\begin{itemize}
  \item \textbf{P2VAD with Non-Identifiable Elements (NIE)} represents approaches that fundamentally alter the data acquisition process to avoid capturing or storing privacy-sensitive information from the outset. Non-identifiable elements refer to video data representations that have been processed, encoded, or captured in ways that obscure or eliminate personally identifiable information while preserving motion and behavioral patterns necessary for anomaly detection. These methods operate on the principle that preventing the initial capture of sensitive data is more secure than attempting to protect such data after acquisition. NIE techniques include encrypted video coding that compresses RGB sequences into indistinguishable binary streams, deployment of Non-Visible Light Cameras (NVLCs) that capture thermal or depth information without revealing facial features, and application of object detection models to mask human-related regions in RGB sequences during preprocessing.
  \item \textbf{P2VAD with Desensitized Intermediate Modalities (DIM)} focuses on the model learning stage by extracting privacy-neutral intermediate representations from video data that capture essential motion and behavioral information while discarding appearance details. Desensitized intermediate modalities are data representations derived from RGB video sequences that preserve temporal and spatial patterns relevant to anomaly detection while removing identifiable visual characteristics. These modalities leverage the insight that most real-world anomalies are motion-related rather than appearance-related, enabling effective detection through representations such as human skeleton keypoints that capture body pose and movement without revealing facial features, clothing, or other identifying characteristics. Optical flow representations that encode pixel-level motion vectors while discarding color and texture information also fall into this category.
  \item \textbf{Edge-Cloud Intelligence (ECI) Empowered P2VAD} addresses privacy risks that emerge during the system application stage, particularly in distributed computing environments where video data or derived features must be transmitted, stored, and processed across multiple devices, edge nodes, and cloud servers. These methods recognize that even when initial data acquisition employs privacy-preserving techniques, additional vulnerabilities can arise from data transmission, collaborative processing, and distributed model training. ECI approaches employ cryptographic techniques such as homomorphic encryption and differential privacy, federated learning protocols that enable collaborative model training without centralizing raw data, and edge-cloud collaboration architectures that process sensitive information locally while leveraging cloud resources for computational scaling.

\end{itemize}

NIE methods can be further categorized based on the format and characteristics of the acquired video data. Encrypted Compressed Video-based approaches utilize standard video compression formats like H.264 or H.265 to transform RGB sequences into compressed bitstreams that appear as indistinguishable binary data to human observers while preserving sufficient information for machine learning algorithms to extract behavioral patterns. Sensitive Region Masking/Synthetic Video-based methods employ pre-trained computer vision models to identify and obscure human figures or other sensitive objects in video frames, or alternatively generate synthetic datasets using avatars and simulated environments that replicate anomalous behaviors without involving real individuals. NVLCs Video-based approaches deploy specialized camera hardware such as infrared thermal cameras or depth sensors that capture environmental and motion information through non-visible light spectra, naturally avoiding the collection of detailed appearance information.

P2VAD methods with DIM, while utilizing RGB sequences during initial capture, focus on extracting intermediate representations that eliminate privacy-sensitive appearance information during the model learning phase. Skeleton-based DIM methods extract human pose keypoints that represent joint positions and limb orientations, enabling the detection of behavioral anomalies through motion pattern analysis while completely discarding facial features, clothing details, and background information. These approaches demonstrate particular effectiveness in human-centric anomaly detection tasks and exhibit improved robustness against illumination changes, weather conditions, and camera viewpoint variations compared to RGB-based methods. Skeleton-based approaches can be further classified into four methodological categories: agent task-based methods that learn normality through reconstruction or prediction tasks, clustering approaches that group motion patterns and identify outliers, distribution modeling techniques that capture statistical properties of normal behaviors, and autoregressive methods that learn temporal dependencies in motion sequences.

ECI-enabled methods address the growing complexity of modern surveillance systems that increasingly rely on distributed computing architectures spanning edge devices, intermediate processing nodes, and cloud-based analytics platforms. These approaches recognize that privacy protection must extend beyond individual data processing to encompass the entire system ecosystem, including data transmission protocols, distributed storage mechanisms, and collaborative learning frameworks. Privacy Computing-based P2VAD employs cryptographic techniques to enable computation on encrypted data without exposing sensitive information. Edge-cloud collaboration VAD distributes processing tasks strategically between local edge devices that handle sensitive data and cloud resources that perform computationally intensive analytics on privacy-protected features. Federated VAD enables multiple surveillance systems or organizations to collaboratively improve anomaly detection models through distributed learning protocols that share model parameters rather than raw data, ensuring that sensitive video information remains localized while benefiting from collective intelligence.

\subsection{Contribution Summary}

The main contributions of this article are as follows:
\begin{itemize}
\item To the best of our knowledge, this is the first survey that focuses on privacy-preserving video anomaly detection. We propose a P2VAD taxonomy comprising three major categories with eight subcategories, systematically summarizing the recent advances in P2VAD across data acquisition, model learning, and system applications, while identifying potential research directions.
\item We present the core challenges, assumptions, and optimization strategies of various P2VAD methods. Additionally, we provide research resources, including available code, public datasets, and key literature, hosted in a GitHub repository for future researchers.
\item We empirically summarize the research challenges and opportunities in P2VAD, offering insights into its development trends in light of evolving AI technologies and real-world demands. Our aim is to guide and standardize future work in this field.
\end{itemize}

The remainder of this article is organized as follows: Section~\ref{sec2} introduces the fundamentals of P2VAD, including methods for acquiring and processing non-identifiable video data, key research areas in deep VAD, and the basics of emerging privacy-preserving techniques such as privacy computing and federated learning. Section~\ref{sec3} discusses non-identifiable element-based approaches. Section~\ref{sec4} explores methods for video anomaly detection using non-sensitive intermediate modalities, such as skeleton data and optical flow. Section~\ref{sec5} covers the construction of trustworthy P2VAD systems from the perspective of edge-cloud intelligence. Section~\ref{sec6} reviews benchmark datasets and evaluation metrics in the literature. Section~\ref{sec7} outlines the challenges, bottlenecks, and future opportunities for P2VAD research. Finally, Section~\ref{sec8} concludes the article. The collected research resource are available at: \href{https://anonymous.4open.science/r/P2VAD-75AF/}{https://anonymous.4open.science/r/P2VAD-75AF/}.

\section{Foundations of P2VAD}~\label{sec2}
\subsection{Pathways of Conventional Video Anomaly Detection}

\begin{figure}[t]
  \centering
  \includegraphics[width=\linewidth]{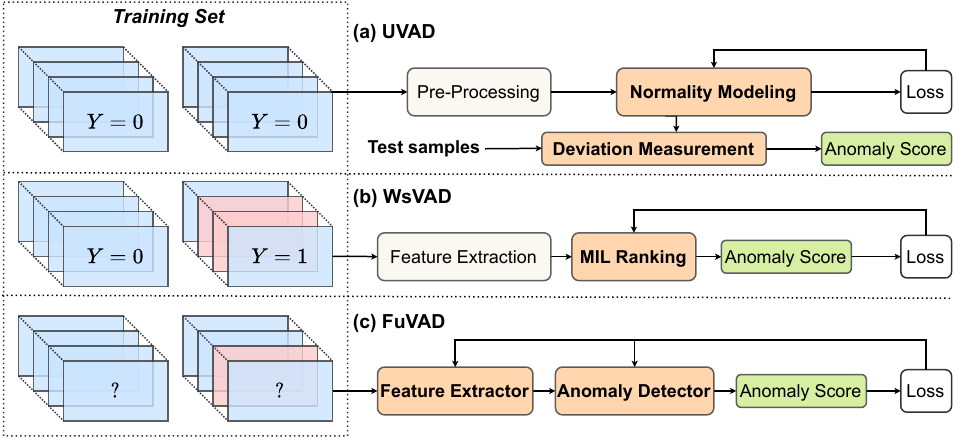}
  \caption{Illustration of the training sets composition and general modeling frameworks of (a) UVAD, (b) WsVAD, and (c) FuVAD. $Y$ denotes the video-level label, and frame-level annotations are unavailable for all pathways.}
  \label{f-pipe}
\vspace{-10pt}
\end{figure}

As mentioned earlier, conventional VAD, without considering data security and privacy protection, is typically categorized into unsupervised, weakly-supervised, and fully-unsupervised approaches based on the supervision signals used, denoted as UVAD, WsVAD, and FuVAD, respectively. The composition and annotations of their training sets as well as the general framework are illustrated in Fig.~\ref{f-pipe}.

\subsubsection{Unsupervised Pathway} 
UVAD trains models using easily collected normal events to capture common spatio-temporal patterns in normal videos through self-supervised learning \cite{cheng2024normality,liu2023learning,cheng2023spatial,xue2024video,huang2022self}. Commonly used supervisory signals include input sequences \cite{FF-AE,memAE} and future frames \cite{FFP,FFPN,Bi-Pre}, corresponding to reconstruction and prediction tasks, respectively. The key assumption is that models trained on normal events cannot represent unseen anomalous samples, which results in significant errors of agent tasks during the test phase. 

Unsupervised VAD follows the anomaly detection community's open-world consensus by treating anomalies as out-of-distribution samples with distinct patterns \cite{pang2021deep}. This approach avoids the need to collect and label low-frequency, unbounded anomalous events and theoretically enables the detection of any possible anomaly in an open environment. As a result, unsupervised methods dominate both P2VAD and deep VAD research \cite{yang2023video,wu2024deep,yan2023feature}. However, unsupervised P2VAD typically utilizes bitstream videos, skeleton data, and other modalities that do not contain identifiable or sensitive information to prevent privacy leakage associated with RGB sequences in deep VAD.

\subsubsection{Weakly-supervised Pathway}
WsVAD was initiated by Sultani et al. \cite{MIR} in 2018 under the Multiple Instance Learning Ranking (MILR) framework, which uses weakly labeled normal and abnormal events to train regression models for computing strong semantic frame-level labels. As shown in Fig.~\ref{f-pipe}(b), the training set of the weakly supervised dataset includes videos with video-level labels, where 0 indicates a regular video and 1 signifies that the video contains abnormal behavior, but the exact temporal location is unknown. 

The core optimization goal of MILR is to train a fully connected network-based regression model, where the highest anomaly score of all instances $\mathcal{V}^i$ in positive bags $\mathcal{B}_a$ (formed by anomalous videos with label 1) is greater than that of the negative bag $\mathcal{B}_n$ from normal videos. This objective is balanced by hyper-parameters $\{\lambda_1, \lambda_2\}$ and formalized as:
\begin{equation}
  \resizebox{0.42\textwidth}{!}{$
  \begin{aligned}
    O\left(\mathcal{B}_a, \mathcal{B}_n\right) & =\bm{\min} \max \left(0,1-\max _{i \in \mathcal{B}_a} r\left(\mathcal{V}_a^i\right)+\max _{i \in \mathcal{B}_n} r\left(\mathcal{V}_n^i\right)\right)\\ & + \lambda_1 \overbrace{\sum_i^{n-1}\left(r\left(\mathcal{V}_a^i\right)-r\left(\mathcal{V}_a^{i+1}\right)\right)^2}^{C_{\text {sm }}}+\lambda_2 \overbrace{\sum_i^n r\left(\mathcal{V}_a^i\right)}^{C_{\text {sp }}},
  \end{aligned}
  $}
\end{equation}
where $C_{sp}$ and $C_{sm}$ represent sparsity and smoothness constraints, inspired by the infrequent and gradual nature of anomalous events. $\mathcal{V}_a^i$ and $\mathcal{V}_n^i$ are instances from $\mathcal{B}_a$ and $\mathcal{B}_n$.

Since WsVAD incorporates anomalous samples during training, it is typically limited to detecting predefined categories of anomalies \cite{liu2022collaborative,wei2022msaf,liu2022learning,wu2024toward}. However, because the model compares spatio-temporal patterns of normal and anomalous events during learning, it is thought to capture the intrinsic differences between them. WsVAD has been shown to produce more reliable results than UVAD \cite{MIST}, especially since UVAD often suffers from high false alarm rates when handling regular videos with label-independent data offsets \cite{wei2022look,cheng2023configurable,10097052}.

Research into weakly supervised P2VAD is still in its early stages, focusing on desensitizing weakly supervised datasets for privacy preservation. For instance, Boekhoudt et al. \cite{boekhoudt2021hr} focused on human-centric crimes in the UCF-Crime dataset, extracting videos with humans and providing skeleton annotations. They developed the first weakly supervised P2VAD dataset, visualizing human motor behavior through skeletons while eliminating the privacy risks associated with RGB data from the Internet. Previous WsVAD approaches used 3D neural networks (e.g., C3D \cite{C3D}, I3D \cite{I3D}, and TSN \cite{TSN}) pre-trained on action recognition datasets like Kinetics \cite{I3D} to obtain spatio-temporal features from RGB sequences, whereas methods based on HR-Crime \cite{boekhoudt2021hr} typically use graph networks to capture skeleton patterns. Subsequent P2VAD studies can leverage the basic assumptions and optimization strategies of deep WsVAD to perform P2VAD on intermediate modalities without relying on privacy-sensitive appearance information.

\subsubsection{Fully-unsupervised Pathway}
FuVAD, a recent approach that follows the transductive learning paradigm \cite{yang2021generalized}, uses datasets that may contain a small number of anomalous samples for training \cite{GCL,SDOR,Deep-UAD}. It challenges the conventional UVAD paradigm, which requires only normal samples during training and involves an additional data filtering process. Instead, FuVAD aims to model vast real-world videos and learn the anomaly classifier directly, as shown in Fig.~\ref{f-pipe}(c).

Given the pattern distribution differences and the low frequency of anomalous events, FuVAD proposes that it is possible to observe many normal samples and construct a model that learns the distributional relationships of the majority of the data. This pattern is treated as in-distribution, with out-of-distribution samples identified as anomalies. Existing deep FuVAD methods often use autoregression \cite{SDOR} or self-training \cite{GCL} to gradually learn the classification boundary. In P2VAD, researchers \cite{9801677} attempt to replace the RGB sequences used in the original method with data such as skeletons to achieve FuVAD without compromising privacy.

\subsection{Appearance Abstraction and Desensitization in P2VAD}
The privacy risks associated with deep VAD primarily stem from the appearance information in RGB sequences, such as human facial features, clothing colors, textures, and sensitive environmental layouts. To address these risks, P2VAD research focusing on desensitization at the data acquisition and model learning stages employs several methods to obscure identifiable information. These include: (1) modeling compressed bitstream videos that are not directly recognizable to the human eye, which both circumvents appearance information and reduces data transmission costs; (2) employing Object Detection (OD) models to identify and mask human-related pixels; (3) using non-visible video sensors, such as event cameras and infrared cameras, to capture environmental data; (4) applying pose estimation models to extract key points of human skeletons for anomaly detection without appearance information; and (5) utilizing optical flow extraction to capture texture-free, detail-free motion information. This section provides an overview of these techniques and models:

\subsubsection{Encrypted Video Encoding}
The core of video encoding techniques is to store and transmit the original RGB video sequences as compressed bitstreams, which not only reduce data size but also obscure appearance information. Common video coding standards, such as H.264 \cite{wiegand2003overview}, H.265, and High Efficient Video Coding (HEVC) \cite{ohm2012comparison}, achieve compression by removing redundant data, resulting in a bitstream that is not directly recognizable by the human eye. These techniques exploit the similarity between video frames and the human eye's sensitivity to high-frequency information to minimize storage requirements. In addition to lowering storage and transmission costs, this method effectively prevents the leakage of identifiable appearance data, such as facial features or clothing color, by focusing on bitstream features. In the context of VAD, replacing RGB data with bitstreams enables models to learn normal video features in compressed space, thus mitigating privacy risks while significantly improving system processing efficiency \cite{8766853,6890212}. For instance, by analyzing motion vectors \cite{biswas2015anomaly} and other encoded features \cite{4761138} in the bitstream, models can detect anomalous behavior patterns quickly without requiring full access to appearance data.


\subsubsection{Object Detection (OD)}
OD techniques are employed to detect and localize target sensitive objects (e.g., humans) within video frames. The application of OD in VAD extends beyond privacy preservation to encompass AD approaches that analyze object-level behaviors and interactions. Object-centric VAD methods have demonstrated significant effectiveness in capturing anomalous patterns by focusing on individual objects or their relationships within scenes. For instance, Ionescu et al. \cite{ionescu2019object} introduce object-centric convolutional auto-encoders that encode both motion and appearance information, formulating abnormal event detection as a one-versus-rest binary classification problem where normality clusters are separated from dummy anomalies. Similarly, Wang et al. \cite{wang2022video} propose a self-supervised approach that solves decoupled spatio-temporal jigsaw puzzles, treating VAD as a multi-label fine-grained classification problem that captures discriminative appearance and motion features through spatial and temporal puzzle permutations. However, object-centric methods in conventional VAD research inherently raise privacy concerns when processing RGB sequences, as they typically rely on detailed appearance information to distinguish between normal and abnormal object behaviors. This limitation motivates the development of privacy-preserving object detection strategies in P2VAD systems. Commonly used OD algorithms, such as the YOLO family \cite{jiang2022review} (e.g., YOLOv3 \cite{redmon2018yolov3} and YOLOv3-spp \cite{huang2020dc}) and Mask R-CNN \cite{he2017mask}, can be repurposed to quickly recognize human figures in images and protect privacy by masking these pixels. Combined with instance segmentation \cite{minaee2021image}, P2VAD can more precisely mask human-related pixels in videos, minimizing privacy exposure and avoiding unnecessary spatial noise. By transmitting only frames that exclude identifiable information, P2VAD enables effective VAD while preserving privacy.

\subsubsection{Non-visible Light Cameras}
Infrared cameras, depth cameras, and event cameras \cite{gallego2020event} capture videos lacking texture and detail but containing spatial and motion information by sensing light outside the visible spectrum. These cameras generate frames by capturing heat, distance, or temporal variations of objects, thus reflecting spatial and temporal changes in the environment without requiring detailed color or texture information. Since these data typically do not reveal facial or bodily details, they naturally offer privacy protection. For example, infrared cameras \cite{ma2019infrared} detect thermal radiation to generate images, while depth cameras use the Time-of-Flight method to measure distances between objects and the camera. Non-visible data is particularly useful in VAD, especially for identifing motion-based anomalies \cite{gaus2023region}, as it can capture large-scale contours and behavioral changes without collecting sensitive appearance information.

\subsubsection{Pose Estimation}
Pose estimation models extract key points to represent skeletal information, ignoring human appearance details. Typical methods, such as OpenPose \cite{martinez2019openpose} and AlphaPose \cite{fang2022alphapose}, use deep learning networks to identify and localize human body joint points in video data, generating 2D or 3D representations of the human skeleton. This ability to directly capture human movement patterns makes pose estimation highly effective for VAD tasks, especially when anomalies involve human motion. Skeleton data is more robust than RGB frames \cite{noghre2024pheva}, as it avoids issues like lighting and occlusion. Moreover, pose estimation is advantageous for privacy preservation, as skeleton information does not reveal personal appearance. As a result, skeleton-based VAD models are widely used in human activity monitoring, where they learn normal movement patterns to detect anomalies.

\subsubsection{Optical Flow}
Optical flow \cite{dosovitskiy2015flownet,ilg2017flownet} computation is a technique that describes pixel-level motion information in videos, capturing the direction and speed of object movement over time. Since optical flow focuses exclusively on motion, ignoring the appearance details of objects, it offers an effective means of privacy preservation in VAD tasks. In P2VAD, optical flow is used to model motion patterns, particularly in the detection of anomalies characterized by motion, capturing temporal dynamics without relying on RGB data. However, optical flow often needs to be combined with other data (e.g., skeletons \cite{su2023prime} or RGB frames \cite{STM-AE,AMMC-net,wang2023memory}) to enhance temporal modeling accuracy. 

\subsection{Privacy Preservation Mechanisms in P2VAD}

While appearance abstraction and desensitization techniques focus on removing or obscuring identifiable information at the data level, system-level privacy preservation mechanisms provide cryptographic and algorithmic protections during data transmission, storage, and processing phases. These mechanisms are particularly crucial in distributed P2VAD systems where video data or intermediate features must traverse multiple devices, edge nodes, and cloud servers. Understanding these foundational techniques is essential for comprehending the P2VAD frameworks discussed in Sec.~\ref{sec5}.

\subsubsection{Homomorphic Encryption}
Homomorphic encryption enables computation directly on encrypted data without requiring decryption, which has emerged as a fundamental technique for privacy-preserving anomaly detection in cloud environments. In P2VAD applications, this cryptographic approach allows cloud servers to perform anomaly detection algorithms on encrypted video features or motion vectors while maintaining complete data confidentiality. The mathematical foundation ensures that operations performed on encrypted data yield equivalent results to those performed on plaintext data, albeit in encrypted form. When multiple cameras transmit encrypted motion vector data to a central server for cross-scene anomaly analysis, homomorphic encryption enables the computation of aggregated anomaly scores without exposing raw motion information. Cheng et al. \cite{cheng2020securead} demonstrate this principle through their SecureAD framework, which implements additive secret-sharing protocols to distribute encrypted video features across multiple computing nodes. This approach ensures that no single entity can reconstruct the original sensitive information, effectively addressing privacy concerns in collaborative VAD scenarios.

\subsubsection{Differential Privacy}
Differential privacy provides mathematical guarantees against individual data identification by introducing carefully calibrated noise into the computation process. This technique has proven particularly valuable in P2VAD systems where multiple surveillance cameras contribute data for collaborative anomaly detection. The core mechanism involves adding statistical noise to query results or intermediate computations such that the presence or absence of any single video sequence has negligible impact on the final output. Giorgi et al. \cite{giorgi2022privacy} apply differential privacy in edge-cloud VAD architectures, where local edge devices transmit noisy feature vectors to cloud aggregators, preventing the reconstruction of information about specific individuals or locations. The noise injection process is governed by privacy budget parameters that mathematically control the trade-off between privacy protection and analytical utility, as formalized in recent privacy computing research \cite{chen2024federated}.

\subsubsection{Multi-Party Secure Computing (MPC)}
MPC protocols enable multiple participants to jointly compute functions over their combined inputs while maintaining input privacy from each other. In distributed P2VAD scenarios, these cryptographic protocols allow different surveillance systems or organizations to collaboratively train anomaly detection models without sharing their respective video datasets. The underlying protocols ensure that each participant learns only the final computed result, such as model parameters or anomaly scores, without gaining access to other participants' sensitive data. Liu et al. \cite{liu2020toward} identify MPC as a critical component for system-level privacy protection in VAD, particularly relevant for city-wide surveillance systems where different agencies must collaborate on anomaly detection while maintaining strict data sovereignty requirements.

\subsubsection{Federated Learning (FL)}
FL extends the privacy preservation paradigm by enabling distributed model training without centralizing raw data. Rather than collecting video data from multiple sources into a central repository, federated protocols allow each participant to train local models on their private datasets and share only model parameters or gradients. The aggregation process combines these distributed updates into a global model that benefits from collective knowledge while preserving individual privacy. Doshi et al. \cite{10354099} exemplify this approach through their FLVAD framework, which implements Transformer-based local models that process video data on trusted edge devices while participating in global model optimization through secure parameter sharing. Similarly, Al-Dujaili et al. \cite{al2024collaborative} propose the CLAP framework, which employs common knowledge-based data segregation and local feedback to enable collaborative training without direct information exchange between clients, further enhancing privacy protection in federated VAD scenarios.

\begin{table*}[t]
\centering
\caption{Technical Comparison of P2VAD methods with Non-Identifiable Elements.}
\label{tab:nie_technical_comparison}
\resizebox{\textwidth}{!}{%
\begin{tabular}{p{2.5cm}p{1cm}p{2cm}p{3.8cm}p{2.4cm}p{2.2cm}p{2.5cm}}
\toprule
\textbf{Method} & \textbf{Year} & \textbf{Category} & \textbf{Core Contributions} & \textbf{Feature Type} & \textbf{Processing Speed} & \textbf{Privacy Mechanism} \\
\midrule

Kiryati et al. \cite{4761138} & 2008 & Encrypted Videos & Motion feature extraction from macroblock motion vectors & Motion vectors & Real-time & Video compression \\

Biswas \& Babu \cite{6776164} & 2014 & Encrypted Videos & Motion vector amplitudes in H.264/AVC compressed domain & Motion vector amplitudes & $>$150 fps & H.264/AVC compression \\

Li et al. \cite{6890212} & 2014 & Encrypted Videos & Motion Intensity Count (MIC) feature using HEVC & Motion vectors, coding units, PU patterns & 1250 fps & HEVC compression \\

Biswas \& Babu \cite{biswas2015anomaly} & 2015 & Encrypted Videos & Enhanced feature set with motion vector orientation & Motion vector amplitudes + orientation & 90$\times$-250$\times$ speedup & H.264/AVC compression \\

Sparse HOMV \cite{6984003} & 2015 & Encrypted Videos & Histogram of motion vectors with sparse representation & HOMV sparse coefficients & Not specified & H.264 compression \\

Guo et al. \cite{8766853} & 2019 & Encrypted Videos & Parameter estimation in H.264 encrypted bitstreams & Macroblock sizes, partition patterns, MV differences & Not specified & H.264 encryption \\

UBnormal \cite{acsintoae2022ubnormal} & 2022 & Synthesis & Synthetic dataset generation for privacy preservation & Synthetic video data & Not specified & Data synthesis \\

Schneider et al. \cite{schneider2022unsupervised} & 2022 & NVLCs & Unsupervised anomaly detection using depth cameras & Depth information & Not specified & Depth sensors \\

TeD-SPAD \cite{fioresi2023ted} & 2023 & Masking & Object detection-based masking approach & Masked visual features & Not specified & Object detection masking \\

Gaus et al. \cite{gaus2023region} & 2023 & NVLCs & Region-based anomaly detection with infrared imaging & Infrared features & Not specified & Infrared imaging \\

Yan et al. \cite{10447842} & 2024 & Masking & Advanced object detection-based masking & Masked regions & Not specified & Object detection masking \\

\bottomrule
\end{tabular}%
}
\end{table*}

\section{P2VAD with Non-Identifiable Elements}~\label{sec3}

\begin{figure}
  \centering
  \includegraphics[width=\linewidth]{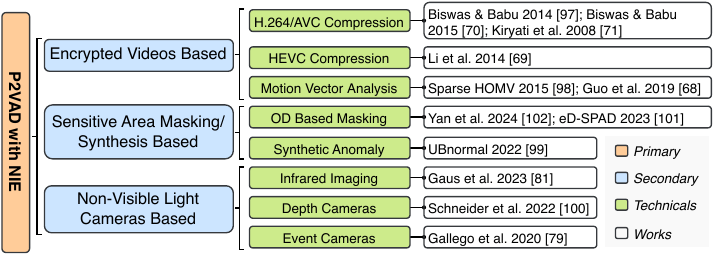}
  \caption{Hierarchical taxonomy and related works of P2VAD with Non-Identifiable Elements (NIE). Note that the listed works include research papers, technical standards, and survey articles that embody related concepts and can inspire corresponding NIE-based P2VAD research directions.}
  \label{f-sec3}
\end{figure}

Non-Identifiable Elements (NIE)-based P2VAD methods aim to prevent privacy and security risks by removing or abstracting identifiable information during data acquisition and preprocessing stages. The desensitized data is then used as input to the anomaly detection model. Based on the level of morphological abstraction, we classify existing NIE methods into three categories: encrypted videos, sensitive area masking/synthesis, and NVLCs videos, as shown in Fig.~\ref{f-sec3}. 
In spite of privacy, such methods reduce data volume and prevent extraneous appearance information from interfering with model learning. However, these approaches have drawbacks, such as reliance on additional processing (e.g., compression, object detection) or the need for specialized acquisition equipment, which can significantly increase the cost of model deployment. The technical comparisons of NIE-based P2VAD methods are summarized in Table~\ref{tab:nie_technical_comparison}.

\subsection{Encrypted Videos-based P2VAD}~\label{sec31}
\begin{table*}[htbp]
  \centering
  \caption{Encrypted Video Coding Techniques for P2VAD.}
  \label{t-evc}
  \begin{tabular}{|p{5cm}|p{5cm}|p{1.5cm}|p{4cm}|}
  \hline
  \textbf{Encrypted Video Coding} & \textbf{Principle} & \textbf{Standard} & \textbf{Encryption Mechanism} \\ \hline
  
  H.264/AVC with Selective Encryption & Encrypts sensitive parts (e.g., motion vectors) while keeping the rest unencrypted. & H.264, AVC & AES-based selective encryption of specific bitstream elements \\ \hline
  
  HEVC with Format-Compliant Encryption & Encrypts specific parts while maintaining standard-compliant streams. & HEVC & Format-compliant encryption of motion vectors or syntax elements \\ \hline
  
  Perceptual Video Encryption& Degrades video quality while preserving basic playability for privacy preservation. & H.264, HEVC & Partial encryption of frames or quantization degradation \\ \hline
  
  Motion-Compensated Encryption & Encrypts motion vectors or residual data to prevent content reconstruction. & H.264, HEVC & AES encryption of motion vectors or residual data \\ \hline
  \end{tabular}
  \end{table*}

Encrypted and compressed videos obscure sensitive appearance information recognizable by humans and filter out low-frequency data, making them inherently suited for detecting anomalous behaviors tied to temporal cues. Additionally, compressed videos require much less storage and bandwidth than raw RGB sequences, reducing transmission costs and enabling real-time VAD. The available encrypted video coding techniques are summarized in Table~\ref{t-evc}. For instance, Biswas and Babu \cite{6776164} found that motion vector amplitudes in the H.264/AVC compressed domain could be effectively used to simulate usual movement patterns and detect anomalies. This method not only addresses privacy concerns but also achieves processing speeds of over 150 frames per second (fps) on the UMN \cite{UMN} and UCSD Peds \cite{ped} datasets, with up to 200$\times$ acceleration. They also introduced a motion pyramid-based hierarchical approach to capture temporal dynamics more efficiently, achieving detection accuracy comparable to state-of-the-art pixel-domain algorithms. In a follow-up study \cite{biswas2015anomaly}, they enhanced the feature set by including motion vector orientation, further improving anomaly detection and achieving 90$\times$ and 250$\times$ speedups on two datasets, respectively. Another method, based on the Histogram of Oriented Motion Vectors (HOMV) within a sparse representation framework \cite{6984003}, utilizes online dictionary learning to represent the sparse behavior of HOMVs. It detects anomalies by analyzing the likelihood of sparse coefficients occurring at specific locations in the H.264 compressed video.

Kiryati et al. \cite{4761138} extracted motion features from the macroblock motion vectors generated during video compression and estimated the statistical distribution of normal events during training. By analyzing improbable-motion eigenvalues, they successfully identified anomalies. Their approach demonstrated strong privacy preservation and real-time performance, making it suitable for video surveillance systems with limited communication and computational resources. Li et al. \cite{6890212} leveraged the correlation between content and coding structure in High-Efficiency Video Coding (HEVC) to develop computationally efficient VAD algorithms. They proposed a Motion Intensity Count (MIC) feature that uses motion vectors, coding units, and prediction unit patterns in HEVC to predict normal motion paths, achieving an average processing speed of 1250 fps. Guo et al. \cite{8766853} further focused on parameter estimation in H.264 encrypted bitstreams to perform VAD without decrypting the video. Their method extracts macroblock sizes, partition patterns, and motion vector differences from encrypted bitstreams, performing anomaly detection after feature fusion.

\subsection{Sensitive Area Masking/Synthesis-based P2VAD}~\label{sec32}

One major concern in VAD research is the lack of public trust, stemming from the fact that surveillance cameras widely deployed in public spaces often capture sensitive personal information such as human faces and body details (including clothing, color, and texture) without consent. When such data is directly used for model training, it not only risks privacy breaches but may also lead to bias in deep learning models, incorporating features related to ethnicity, skin color, and gender. Given that VAD primarily focuses on modeling motion interactions in the temporal domain, appearance information often contributes little to anomaly detection and can introduce sensitivity to irrelevant pixel variations (e.g., changes in lighting, weather, or camera angles). Consequently, a promising P2VAD strategy is to replace or mask sensitive regions in the RGB sequence and use the desensitized data as model input. Researchers have proposed using object detection models and U-Net architectures to mask human information in RGB sequences before modeling video normality. Additionally, synthetic datasets like UBnormal \cite{acsintoae2022ubnormal} leverage virtual data engines to simulate various types of anomalous behaviors, where the virtual humans contain only contour information and lack facial details. As a result, models trained on these datasets focus on motion interactions rather than local appearance details, allowing the use of RGB sequences without identifiable information. This approach ensures the VAD models remain unbiased and more ethically reliable, even though they may still use RGB inputs during real-world applications.

While object detection has been shown to enhance VAD performance, most existing studies \cite{9008722,9626932,9746420} use foreground objects as a semantic supplement to RGB-based normality learning, emphasizing performance in complex scenarios at the cost of privacy preservation. Yan et al. \cite{10447842} proposed a privacy-preserving VAD approach that incorporates an image segmentation mask to safeguard the privacy of human subjects. They introduced a VAD model based on ST-AE and CONV-AE that integrates contextual information while maintaining privacy. Specifically, they utilized Mask-RCNN \cite{he2017mask} to detect targets of interest, generating masks to protect human-related pixels. The model then uses privacy-neutral human profiles and background information as input for normality learning. Similarly, the Temporal Distinctiveness for Self-supervised Privacy-preservation in video Anomaly Detection (TeD-SPAD) method introduced by Fioresi et al. \cite{fioresi2023ted} employs a U-Net to anonymize video frames before performing model training and anomaly detection. The UCF-Crime dataset, used in their experiments, presents unique privacy concerns due to its socially significant crime content, yet its complexity surpasses that of UCSD Ped1 and Ped2 used in \cite{10447842}. In response to these challenges, TeD-SPAD proposes a temporally distinct ternary loss function to efficiently process anonymized videos. Experimental results demonstrate that their approach reduces private attribute predictions by 32.25\%, with only a 3.69\% reduction in frame-level AUC on the UCF-Crime dataset \cite{MIR}.

UBnormal, introduced by Acsintoae et al. \cite{acsintoae2022ubnormal} in 2022, is the first large-scale dataset designed for supervised open-set VAD. While initially developed to address the issue of supervised models failing to recognize open-set samples, UBnormal also mitigates privacy concerns by synthesizing anomalous events using a virtual data engine. The dataset includes various synthetic anomalous behaviors in multiple scenes, where the background is real but the subjects are computer-generated. This synthetic approach enables the development of unbiased VAD models, demonstrating the potential of synthetic datasets for privacy-preserving anomaly detection in P2VAD.

\subsection{Non-Visible Light Cameras-Based P2VAD}
In contrast to the methods discussed in Sec.~\ref{sec31} and Sec.~\ref{sec32}, which primarily process existing RGB sequences, approaches using NVLCs rely on specialized sensing devices such as depth cameras, infrared imagers, and event cameras. These technologies cannot be directly applied to conventional CCTV surveillance systems, and due to the limited availability of such data, research in this area is still in its early stages. However, NVLCs-based VAD systems have shown significant promise in extreme conditions (e.g., dark nights, fog, or high-speed motion scenarios). Their modeling processes typically follow learning frameworks and optimization strategies similar to those used in deep learning-based VAD systems, making them promising candidates for scene-specific P2VAD applications.

Gaus et al. \cite{gaus2023region} proposed a dual approach for anomaly detection in infrared surveillance imagery, utilizing both visual appearance and localized motion properties from optical flow. Their Long-Term infrared (thermal) Imaging (LTD) benchmark validates the effectiveness of this model. This approach to normality learning is conceptually similar to multi-stream unsupervised VAD methods, which use different branches to independently learn appearance and motion features, combining task errors from multiple agents for anomaly detection. Schneider et al. \cite{schneider2022unsupervised} highlighted that depth information allows for the easy extraction of auxiliary data, such as foreground masks, to support scene analysis and aid in anomaly detection. They evaluated the performance of autoencoder-based depth VAD methods on depth video and suggested integrating depth data into the loss function to enhance model performance.

Event cameras \cite{gallego2020event}, which asynchronously measure the luminance change of each pixel to generate an event video stream, offer advantages such as low latency, high speed, and high dynamic range. While no existing work has yet explored the use of event cameras for anomaly detection, studies have demonstrated their potential for characterizing object motion, such as optical flow, using event streams. This type of motion information is crucial for detecting video anomalies, suggesting that event cameras could play a valuable role in future VAD research.

\section{Desensitized Intermediate Modalities P2VAD}~\label{sec4}

\begin{table*}[t]
\centering
\caption{Technical Comparison of DIM-based P2VAD Methods}
\label{tab:dim_technical_comparison}
\resizebox{\textwidth}{!}{%
\begin{tabular}{p{2.5cm}p{1.2cm}p{2cm}p{3.5cm}p{2.5cm}p{2cm}p{2cm}}
\toprule
\textbf{Method} & \textbf{Year} & \textbf{Category} & \textbf{Core Contributions} & \textbf{Feature Type} & \textbf{Learning Framework} & \textbf{Network Architecture} \\
\midrule

Gatt et al. \cite{8868795} & 2019 & Skeleton-Agent & LSTM+CNN autoencoder for normal event representation & Body keypoints & Reconstruction & LSTM+CNN \\

Markovitz et al. \cite{markovitz2020graph} & 2020 & Skeleton-Clustering & Graph Embedded Pose Clustering with GCAE & Human pose maps & Deep clustering & GCAE \\

Rodrigues et al. \cite{rodrigues2020multi} & 2020 & Skeleton-Agent & Multi-timescale model for pose trajectory prediction & Pose trajectories & Prediction & Multi-timescale model \\

Temuroglu et al. \cite{temuroglu2020occlusion} & 2020 & Skeleton-Agent & Occlusion-Aware Skeleton Trajectory Representation & Skeleton trajectories & Reconstruction & Autoencoder \\

Cui et al. \cite{cui2021prototype} & 2021 & Skeleton-Clustering & Prototype generation module with SST-GCN & Skeleton features & Deep clustering & SST-GCN \\

HSTGCNN \cite{9645572} & 2021 & Skeleton-Agent & Hierarchical high/low-level graph representations & Graph representations & Prediction & ST-GCN \\

Liu et al. \cite{liu2021self} & 2021 & Skeleton-Clustering & Spatial-temporal self-attention with graph convolution & Joint-level information & Deep clustering & SAA-Graph \\

Luo et al. \cite{luo2021normal} & 2021 & Skeleton-Agent & ST-GCN for future skeleton prediction & Graph connectivity & Prediction & ST-GCN \\

Suzuki et al. \cite{9385618} & 2021 & Skeleton-Agent & Autoencoder for children's gross motor skills detection & Movement patterns & Reconstruction & Autoencoder \\

Fan et al. \cite{fan2022video} & 2022 & Skeleton-Agent & CycleGAN for enhanced feature extraction accuracy & Enhanced features & Reconstruction & CycleGAN \\

Jiang et al. \cite{jiang2022deep} & 2022 & Skeleton-Agent & Deep NN for pedestrian behavior detection & Keypoint trajectories & Trajectory tracking & Deep NN \\

Li et al. \cite{9801677} & 2022 & Skeleton-Autoregressive & Self-Trained SGCN with iForest initialization & Skeleton features & Self-trained regression & SGCN \\

Tani \& Shibata \cite{tani2022frame} & 2022 & Skeleton-Distribution & Frame-wise AGCN with Gaussian distribution modeling & Action-ness features & Distribution modeling & AGCN \\

Yang et al. \cite{9746420} & 2022 & Skeleton-Clustering & Dual-branch model for humans and objects & Skeleton+object features & Clustering & Dual-branch \\

Chen et al. \cite{chen2023multiscale} & 2023 & Skeleton-Clustering & Multiscale attention-based adjacency matrices & Hierarchical graph features & Deep clustering & MSTA-GCN \\

Javed et al. \cite{javed2023learning} & 2023 & Skeleton-Clustering & Decoder-free direct clustering of skeleton features & Graph convolution features & Direct clustering & GCN \\

Sato et al. \cite{sato2023prompt} & 2023 & Skeleton-Distribution & User cue-guided zero-shot learning framework & User cue embeddings & Distribution modeling & Zero-shot framework \\

Song et al. \cite{song2023analysis} & 2023 & Skeleton-Agent & GAN model for global/local motion features & Motion features & Adversarial learning & GAN \\

Su et al. \cite{su2023prime} & 2023 & Optical Flow & Prime framework with optical flow and skeleton fusion & Optical flow + skeleton & Multi-modal fusion & NMS strategy \\

Yan et al. \cite{10032201} & 2023 & Skeleton-Clustering & Deep memory clustering with real-time updates & Memory features & Memory clustering & GC-Autoencoder \\

\bottomrule
\end{tabular}%
}
\end{table*}

RGB video sequences contain rich appearance and motion semantics but also include a substantial amount of task-irrelevant redundant data. As a result, these sequences are typically processed into more information-dense intermediate modalities, such as facial keypoints, skeletal gestures, and optical flow. The latter two modalities are particularly useful for detecting human-related abnormal behaviors, as they effectively capture motion cues. Many existing skeleton-based VAD methods share a similar modeling process with deep VAD methods. They rely on easily collectible routine events to learn video normality and detect anomalies by measuring deviations between input samples and the learned model during the testing phase. Therefore, skeleton-based VAD methods are classified into four categories based on the normality learning framework: agent task, clustering, distributional modeling, and autoregression.

\begin{figure}
  \centering
  \includegraphics[width=\linewidth]{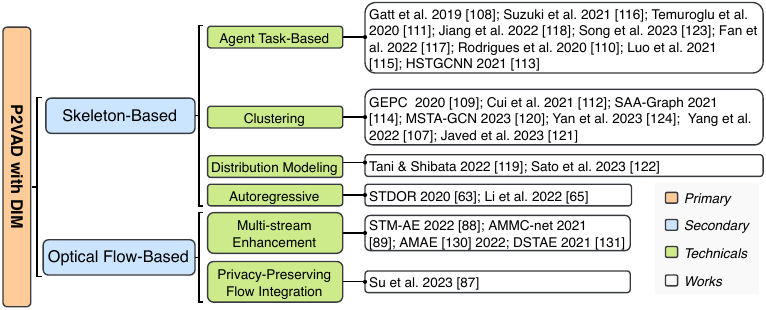}
  \caption{Hierarchical taxonomy and related works of P2VAD with Desensitized Intermediate Modalitie (DIM).}
  \label{f-sec4}
\end{figure}

Newer techniques, such as multi-task learning \cite{sun2023hierarchical} and causality learning \cite{sun2023learning,lv2023unbiased,li2024causaltad,10962292}, have been explored in RGB-based VAD models, allowing for more specialized categories. This progress is expected to guide future skeleton-based VAD research. In contrast, optical flow removes scene appearance information and is less information-dense than skeletal data. Therefore, it is often used as complementary semantics to RGB sequences and is not typically applied as a standalone VAD approach. Studies have shown that such intermediate modalities, which focus on temporal information, not only mitigate privacy concerns but also avoid enrollment sensitivities similar to those in RGB-based methods. In real-world scenarios, factors such as scene changes, camera angles, and weather conditions cause significant pixel-level variations in color images, leading to high false positive rates in unsupervised VAD methods that rely solely on normal video modeling. Scene-robust VAD has thus become a long-standing challenge. The Desensitized Intermediate Modalities (DIM) approach, which focuses on motion information while ignoring background scenes, allows VAD to concentrate on learning scene-robust video normality. The hierarchical taxonomy of P2VAD methods with DIM is shown in Fig.~\ref{f-sec4}, while the technical comparisons of each method are summarized in Table~\ref{tab:dim_technical_comparison}.

\subsection{Skeleton-based P2VAD}

\subsubsection{Agent Task-Based Methods}
In anomaly detection, the open-world assumption acknowledges that anomalous events are diverse and unbounded, making it nearly impossible to collect all potential anomalies for modeling. As such, the task is often treated as out-of-distribution detection, where models learn in-distribution patterns from normal samples, and samples falling outside these patterns are classified as anomalies. Agent task-based approaches are dominant because of their straightforward assumptions and excellent performance on video data. These methods train models to perform self-supervised tasks such as reconstruction or prediction to model normal events. Anomalies are detected during testing by measuring agent task errors.

\begin{figure}[t]
  \centering
  \includegraphics[width=\linewidth]{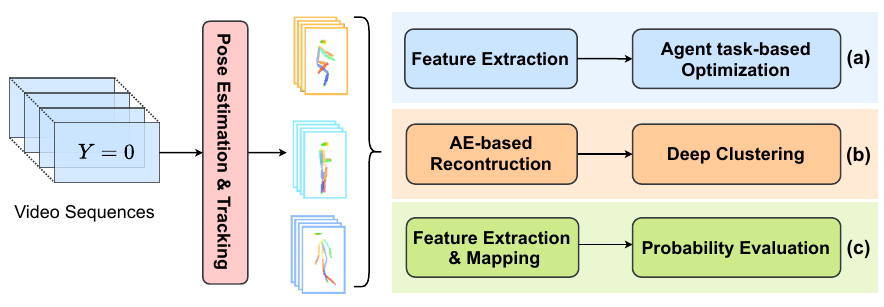}
  \caption{The general pipline of unsupervised skeleton VAD methods based on \textbf{(a)} agent task (Upper), \textbf{(b)} clustering (Middle), and \textbf{(c)} distribution (Lower).}
  \label{f-sk}
\vspace{-10pt}
\end{figure}

Both deep VAD methods using RGB sequences and skeleton-based P2VAD methods follow this approach, although the former typically employs convolutional neural networks or Transformers to extract spatio-temporal features, while the latter models skeletal keypoint sequences, as illustrated in Fig.~\ref{f-sk}(a). For instance, Gatt et al.~\cite{8868795} used PoseNet and OpenPose to detect individuals in frames and extract body keypoints, constructing an autoencoder with LSTM and CNN units to learn representations of normal events. Similarly, Suzuki et al.~\cite{9385618} focused on children's gross motor skills and proposed an autoencoder-based reconstruction framework to detect abnormal movements. These methods, like deep VAD, rely on reconstructing regular event skeletons to model normality.

Temuroglu et al.~\cite{temuroglu2020occlusion} addressed skeleton occlusion in crowd anomaly detection, proposing an Occlusion-Aware Skeleton Trajectory Representation to compare skeleton maps before and after occlusion and embed the output into an autoencoder for robust anomaly detection. Jiang et al.~\cite{jiang2022deep} developed a deep neural network-based method for detecting abnormal pedestrian behaviors (e.g., crouching, wandering) in level-crossing videos, using a skeleton detection algorithm to track keypoint trajectories. In extended work, Song et al.~\cite{song2023analysis} introduced a GAN model to capture global and local motion features of skeletons, using a generator to learn regular patterns and a discriminator for anomaly detection. Similarly, Fan et al.~\cite{fan2022video} employed CycleGAN \cite{zhu2017unpaired} to enhance feature extraction accuracy in skeleton-based VAD.

While reconstruction-based tasks are simpler, they may lead neural networks to overlook abstract semantic features. In contrast, prediction tasks require the model to reason about the spatio-temporal evolution of input sequences, proving more effective in VAD. Inspired by this, researchers have developed skeleton-based VAD using prediction tasks. For example, Rodrigues et al.~\cite{rodrigues2020multi} proposed a multi-timescale model for predicting pose trajectories, enabling anomaly detection at varying timescales. Luo et al.~\cite{luo2021normal} employed spatio-temporal graph convolutional networks to predict future skeletons and identify anomalous behaviors based on graph connectivity. Hierarchical Spatio-Temporal Graph Convolutional Neural Networks (HSTGCNN)~\cite{9645572} further distinguished between high- and low-level graph representations, achieving excellent performance on benchmark datasets. 

\subsubsection{Clustering Methods}
Unlike agent task-based methods, which use task errors to detect anomalies, clustering-based methods group low-dimensional skeleton features into clusters, with samples far from any normal clusters considered anomalies. The general pipeline of clustering-based skeleton VAD is illustrated in Fig.~\ref{f-sk}(b). For example, Graph Embedded Pose Clustering~\cite{markovitz2020graph} uses GCAE to map human pose maps into latent space, applying deep clustering for fine-grained detection of anomalous behaviors. Similarly, Cui et al.~\cite{cui2021prototype} proposed a skeleton VAD model with a prototype generation module embedded between a Spatio-Temporal Graph Convolutional Network (SST-GCN) and a deep clustering layer. 

Spatial-temporal self-attention augmented graph convolutional networks (SAA-Graph)~\cite{liu2021self} combine improved spatial graph convolution with Transformer Self-Attention to capture joint-level information and cluster it using deep embedded clustering. The anomaly score of test samples is computed solely based on clustering, which reduces inference costs. Multiscale Spatiotemporal Attention Graph Convolutional Networks (MSTA-GCN)~\cite{chen2023multiscale} apply attention-based adjacency matrices to capture hierarchical graph representations. Yan et al.~\cite{10032201} introduced deep memory clustering for real-time updates of pseudo-labels and network parameters using a graph convolutional autoencoder. Additionally, Yang et al.~\cite{9746420} proposed a dual-branch model for detecting both anomalous human behaviors and objects, where skeletons are clustered to compute anomaly scores. Javed et al.~\cite{javed2023learning} removed the decoder structure commonly used in clustering-based methods, directly clustering skeleton features extracted by graph convolution operators to enhance model stability.

\subsubsection{Distribution Modeling Methods}
Tani and Shibata \cite{tani2022frame} attempted to model the distributional relationships of skeleton information for normal events and perform anomaly detection using probabilistic methods, as show in Fig.~\ref{f-sk}(c). Specifically, they first use an Adaptive Graph Convolutional Network (AGCN), pre-trained on the Kinetics dataset, to obtain the "action-ness" of each frame and then select specific frames for further training of frame-wise AGCN models. The trained frame-wise AGCN with fixed parameters computes representations of normal events and constructs multivariate Gaussian distribution models. Anomaly scores for test samples are then obtained by calculating their Mahalanobis distances within this distribution. Similarly, Sato et al. \cite{sato2023prompt} attempt to detect video anomalies through a distribution modeling approach by proposing a user cue-guided zero-shot learning framework. This method models the distribution of skeleton features during training and uses it in the inference phase to estimate anomaly scores. To reduce false alarms, the authors integrate the similarity score between user cue embeddings and the skeleton features aligned in the shared space into the final anomaly score, indirectly enhancing the detection of normal actions.

\subsubsection{Autoregressive Methods}
Whether by using surrogate tasks to indirectly detect anomalies or by clustering and distribution modeling to directly characterize the pattern boundaries of normal events, these methods fall under the transductive learning paradigm, which requires that the training set contain only normal samples. While feasible for small datasets or videos from low-incidence anomalous scenarios, accurately filtering all potential anomalies from large-scale surveillance videos is both time-consuming and labor-intensive. To address this, recent research has shifted towards fully unsupervised VAD, directly learning anomaly detection models from training samples that contain a small number of anomalies (determined by the low frequency of anomalies rather than explicit labeling). In the realm of deep VAD using RGB image sequences, Pang et al. \cite{SDOR} proposed the Self-Trained Deep Ordinal Regression (STDOR) framework to progressively identify a small number of anomalous events that differ from the large number of samples displaying significant spatio-temporal patterns, using iterative learning. Furthermore, Li et al. \cite{9801677} addressed the privacy preservation challenge in FuVAD by employing pose estimation algorithms to extract skeletons from RGB sequences without revealing sensitive appearance information. They then developed a Self-Trained Spatial Graph Convolutional Network (SGCN) for P2VAD. The approach first uses iForest \cite{zhao2018iforest} to roughly distinguish between possible abnormal and normal skeletons. An anomaly scoring module, consisting of an SGCN and a fully connected layer, then computes the anomaly score, which serves as the basis for discriminating possible positive and negative samples in subsequent iterations, overseeing the model's continuous optimization via self-trained regression.

\subsection{Optical Flow-based P2VAD}
Optical flow and skeleton data, both derived from RGB sequences, significantly enhance the credibility of VAD studies by discarding sensitive appearance information, such as facial details and clothing texture. Additionally, optical flow reduces visual information redundancy by focusing on motion cues directly associated with abnormal behavior, which reduces data size and eases model training complexity. However, optical flow captures pixel-level motion, which has a lower information density compared to skeleton data and is more sensitive to image noise. Moreover, it struggles to handle overlapping motion trajectories in complex scenes, limiting its effectiveness in crowd anomaly detection. Consequently, optical flow is typically used as complementary semantics to enhance normality learning \cite{AMMC-net,AMAE,DSTAE}, rather than being employed alone in VAD.

Deep multi-stream UVAD models \cite{STM-AE,wang2023memory,zhao2022lgn,CDD-AE+} often construct separate branches to learn motion patterns from optical flow streams, which are then combined with RGB-based branches to improve the model's capacity to capture spatio-temporal normality. Compared to single-stream methods that only use RGB sequences, the introduction of optical flow allows the model to focus on temporal dynamics while ignoring background noise. However, these methods still rely on RGB sequences as input, posing privacy concerns. In response, Su et al. \cite{su2023prime} analyzed pixel motion using optical flow between consecutive frames, which they then integrated with skeletal joint positions for anomaly detection. Their proposed privacy-preserving video anomaly detection framework, named Prime, leverages optical flow and skeleton data as inputs, employing a Non-Minimal Suppression (NMS) strategy to adaptively highlight the consistency of anomalies between the two modalities.

\section{Edge-Cloud Intelligence Empowered P2VAD}~\label{sec5}

\begin{table*}[htbp]
\centering
\caption{Technical Comparison of ECI empowered P2VAD Methods}
\label{tab:eci_technical_comparison}
\resizebox{\textwidth}{!}{%
\begin{tabular}{p{2.5cm}p{1.2cm}p{2cm}p{4cm}p{2.5cm}p{2.5cm}p{2cm}}
\toprule
\textbf{Method} & \textbf{Year} & \textbf{Category} & \textbf{Core Contributions} & \textbf{Privacy Technique} & \textbf{System Architecture} & \textbf{Key Features} \\
\midrule

Liu et al. \cite{liu2020toward} & 2020 & Privacy Computing & System-level data security and privacy preservation & Multi-Party Secure Computing & Distributed system & MPC collaboration \\

Cheng et al. \cite{cheng2020securead} & 2020 & Privacy Computing & SecureAD framework with additive secret-sharing & Homomorphic encryption & Deep neural networks & Secret-sharing protocols \\

Giorgi et al. \cite{giorgi2022privacy} & 2022 & Edge-Cloud & Differential privacy in end-edge-cloud data exchange & Differential privacy & Edge-cloud architecture & Private feature vectors \\

Wen et al. \cite{wen2023survey} & 2023 & Federated VAD & Comprehensive survey of federated learning approaches & Distributed learning & Federated architecture & Survey paper \\

Doshi et al. \cite{10354099} & 2023 & Federated VAD & FLVAD: Transformer-based federated anomaly detection & Federated learning & Transformer architecture & Local+global models \\

Al-Dujaili et al. \cite{al2024collaborative} & 2024 & Federated VAD & CLAP: Collaborative Learning with Privacy & Federated learning & Collaborative framework & Pseudo-labeling \\

Chen et al. \cite{chen2024federated} & 2024 & Privacy Computing & Federated learning with differential privacy integration & Differential privacy & Federated system & Noise-based protection \\

Liu et al. \cite{liu2024networking} & 2024 & Edge-Cloud & Networking systems framework for VAD & Secure transmission & Edge-cloud integration & Hardware+algorithm \\

Wang et al. \cite{wang2024end} & 2024 & Edge-Cloud & End-to-end distributed architecture & Edge reasoning & End-edge-cloud & Task distribution \\

\bottomrule
\end{tabular}%
}
\end{table*}

With the growing adoption of video IoT systems in smart cities and the expansion of online video platforms, data transmission and processing in VAD systems have transitioned from centralized local devices to distributed hybrid architectures that combine edge and cloud computing. Edge-cloud intelligence-enabled P2VAD research focuses on leveraging privacy computing and distributed privacy-preserving techniques to enhance the security of VAD systems across end devices, edge nodes, and the cloud. Unlike the approaches presented in Sections~\ref{sec3} and~\ref{sec4}, which focus on privacy risks during data acquisition and preprocessing by removing identifiable appearance information, Edge-Cloud Intelligence (ECI) empowered VAD emphasizes system-level data security and privacy preservation \cite{liu2020toward}. Although localized VAD models may still use RGB sequences as inputs, these computational processes typically occur on user-trusted devices, such as smartphones or PCs. Instead of collecting raw data, data centers and cloud servers process encrypted information or intermediate features, ensuring that even in the event of a system breach, sensitive information remains secure. Furthermore, the distributed processing of surveillance videos from different scenes prevents distribution bias caused by simple data aggregation, allowing cloud models to perform cross-scene anomaly detection.

\begin{figure}
  \centering
  \includegraphics[width=\linewidth]{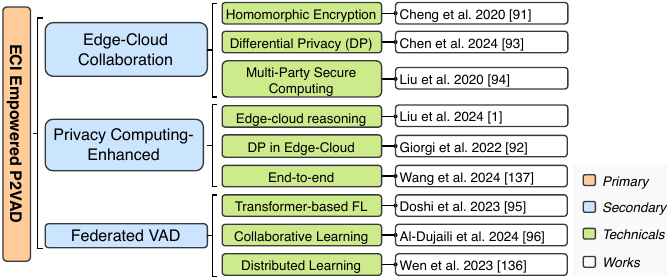}
  \caption{Hierarchical taxonomy and related works of End-Cloud Intelligence (ECI) Empowered P2VAD.}
  \label{f-sec5}
\end{figure}

VAD systems for large-scale applications handle massive amounts of video data, which typically need to be transmitted and processed across various edge devices, servers, and data centers. Without effective privacy protection mechanisms, the exchange and integration of information in edge-cloud systems pose significant risks of data leakage and misuse. Therefore, system-level privacy protection not only safeguards user data but also strengthens societal trust in such systems. This chapter explores three major directions: (1) Privacy Computing-enhanced VAD, which reduces the risk of data leakage through privacy computing techniques; (2) Edge-Cloud Collaboration VAD, which focuses on data security and performance optimization during collaboration between end devices and the cloud; and (3) Federated VAD, which explores privacy-secure VAD model training via distributed learning without sharing raw data. These approaches complement one another and collectively improve system-level privacy security. The taxonomy and summary of NIE empowered P2VAD methods are provided in Fig.~\ref{f-sec5} and Table~\ref{tab:eci_technical_comparison}, respectively.

\subsection{Privacy Computing-Enhanced VAD}
Privacy computing \cite{chen2024federated} ensures that data is transmitted and processed without exposing sensitive information, especially when video data is distributed across different devices and servers. Key techniques include: (1) Homomorphic encryption, which supports direct computation on encrypted data and is commonly used to ensure data privacy in cloud computing for VAD systems. For instance, data from multiple cameras can be encrypted homomorphically, allowing anomalous behavior analysis to be performed on the encrypted data; (2) Differential privacy, which ensures that individual data points have a negligible effect on the overall analysis by adding noise to the data or query results. This technique is particularly useful for analyzing real-time surveillance data in VAD systems, preventing sensitive information leakage; and (3) Multi-Party Secure Computing (MPC), which allows multiple participants to compute the result of a function collaboratively without sharing their respective data. In VAD systems, MPC enables devices, such as edge cameras, to collaborate in detecting anomalies without exposing their data.

Significant progress has been made in privacy computing-enhanced P2VAD, attracting attention from the anomaly detection and AI security communities. For instance, homomorphic encryption allows the detection of anomalous behaviors in video frames without decrypting the video itself. Differential privacy techniques facilitate aggregated video data analysis while preserving privacy. Cheng et al. \cite{cheng2020securead} propose a secure video anomaly detection framework called SecureAD, which safeguards the security of deep neural networks through additive secret-sharing protocols. Additionally, the authors introduce a fine-grained Bloom filter-based access control policy to authenticate legitimate users without compromising the privacy of original personal attributes.

The application of privacy computing techniques broadens the potential use cases of VAD systems and enhances user trust, particularly in multi-party data collaboration scenarios. These techniques ensure data security during transmission and processing, reduce the risk of privacy breaches, and can be seamlessly integrated with existing deep VAD algorithms to maintain performance while improving privacy protection.

\subsection{Edge-Cloud Collaboration VAD}
Edge-Cloud Collaboration VAD is a distributed architecture that strategically divides computing tasks between edge devices and the cloud, aiming to maximize the use of computing resources and alleviate users' privacy and security concerns about data centers \cite{wang2024end}. In this framework, video data captured from real-world scenes can be initially processed on edge devices close to the data source, and only essential features or privacy-protected data are sent to the cloud for further analysis. Since only user-trusted local devices access raw data containing identifiable information, the end-edge-cloud data exchange process and global model training can avoid the risk of data theft and misuse. Existing research in this area is typically categorized into three dimensions: (1) Edge reasoning: smartphones and PCs deployed on the end-side utilize local computing power to preprocess video streams and detect simple anomalies in real-time. Lightweight deep VAD models are often used, reducing data transmission and minimizing privacy exposure; (2) Cloud aggregation: the cloud collects intermediate outputs from the edge, such as hidden features or decision vectors devoid of identifiable information, to train cross-scene models for detecting complex, long-duration anomalies; and (3) Secure data transmission: data is encrypted during transmission across devices, ensuring that sensitive information is not leaked.

In their work on Networking Systems for VAD, Liu et al. \cite{liu2024networking} propose an edge-cloud system framework that integrates hardware and algorithms to provide a secure and efficient communication and computation architecture for deep VADs. The framework includes a deployment platform, a data pipeline, and a computation container for data collection, transmission, and anomaly detection. The data pipeline incorporates various encryption protocols to prevent privacy breaches. Giorgi et al. \cite{giorgi2022privacy} introduce differential privacy into the end-edge-cloud data exchange process to address potential data security risks between clients and data centers. They run the feature extraction process locally, transmitting only differentially private hidden vectors to a central server. A cloud-based decoder reconstructs anonymized versions of the original data frames and computes spatial, temporal, and contextual features for anomaly detection via supervised learning.

\subsection{Federated VAD}

\begin{figure}[t]
  \centering
  \includegraphics[width=\linewidth]{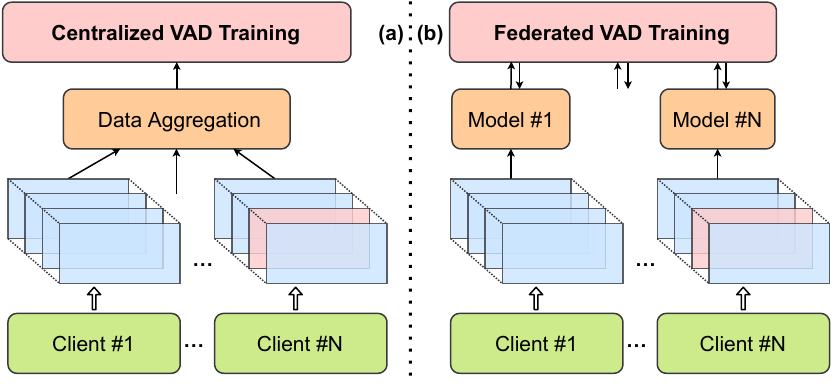}
  \caption{Pippeline of (a) Data aggregation-based centralized VAD and (b) Federated VAD with parameter sharing for privacy-preservation \cite{al2024collaborative}.}
  \label{f-fe}
\vspace{-10pt}
\end{figure}

Federated Learning (FL) \cite{wen2023survey} is a distributed machine learning approach that enables multiple devices to collaboratively train a global model while preserving privacy by not sharing raw data, as illustrated in Fig.~\ref{f-fe}. This technique is widely used in applications such as autonomous driving and robotics. As public concerns about privacy in machine learning applications, including VAD, continue to grow, researchers have increasingly introduced federated learning into VAD, proposing federated VAD solutions to address issues such as skewed data distributions in heterogeneous devices. Doshi et al. \cite{10354099} propose a Transformer-based privacy-preserving video anomaly detection and behavior recognition framework. The FLVAD architecture they introduce consists of multiple local models and a global model, all implemented via Transformer, allowing local data processing while the cloud aggregates the local models trained on different datasets. Collaborative Learning of Anomalies with Privacy (CLAP) \cite{al2024collaborative} addresses large-scale VAD applications by avoiding direct information exchange or data aggregation between clients, protecting user privacy. CLAP employs common knowledge-based data segregation and local feedback to improve pseudo-labeling and enable collaborative training.

Federated learning techniques are expected to mitigate the data privacy risks associated with centralized model training and inference in city-wide surveillance systems. For example, multiple surveillance systems from different locations (e.g., neighborhoods, parks, hospitals, stations) in a city can use federated learning to jointly train anomaly detection models without sharing their respective video data.

\section{Evaluation Benchmarks and Metrics}~\label{sec6}
\subsection{Benchmark Datasets}
Privacy-Preserving Video Anomaly Detection (P2VAD) has evolved as a subset of traditional anomalous behavior detection, driven by increasing concerns over information security. Most P2VAD studies continue to use established RGB datasets for performance evaluation, often preprocessing or reorganizing them to remove sensitive appearance information. For example, encrypted video-based methods use encrypted, compressed RGB videos, while skeleton-based VAD models focus on human-centric video datasets, using pose estimation algorithms to model human skeletons and preserve privacy. Additionally, system-level VAD research emphasizes security during the transmission and distribution of data, with local clients typically performing anomaly detection on models trained with RGB sequences. 

We classify the available datasets into two categories: 1) \textit{generalized datasets}, which are traditional RGB videos widely used in VAD research and can be adapted for various tasks after preprocessing; and 2) \textit{specialized datasets}, which are stripped of identifiable and sensitive information and are only suitable for specific P2VAD tasks. Table 1 summarizes key statistics for popular generalized datasets, which are further divided into unsupervised and weakly supervised types based on the availability of labels.

\subsubsection{Generalized VAD Datasets}

\begin{table}[]
  \centering
  \caption{Statistical information of the generalized VAD datasets.}
  \label{t-data}
  \begin{threeparttable}
  \resizebox{\linewidth}{!}{
  \begin{tabular}{@{}lccccc@{}}
  \toprule
  \multicolumn{1}{c}{\multirow{2}{*}{\textbf{Dataset}}} & \multicolumn{2}{c}{\textbf{\#Videos}} & \multirow{2}{*}{\textbf{\#Scenes}} & \multirow{2}{*}{\textbf{\#Classes}} & \multirow{2}{*}{\textbf{\#Anomalies}} \\ \cmidrule(lr){2-3}
  \multicolumn{1}{c}{}  & Training           & Testing          &    &     &       \\ \midrule
  Subway Entrance \cite{Subway}       & -   & - & 1  & 5   & 51    \\
  Subway Exit \cite{Subway}            &  -  & - & 1  & 3   & 14    \\
  UMN$^\dagger$ \cite{UMN}  &  -  & - & 3  & 3   & 11    \\
  CUHK Avenue \cite{avenue}          & 16 & 21  & 1  & 5   & 77    \\
  UCSD Ped1 \cite{ped} & 34 & 36  & 1  & 5   & 61    \\
  UCSD Ped2 \cite{ped} & 16 & 12  & 1  & 5   & 21    \\
  ShanghaiTech \cite{FFP}          & -   &  & 13 & 11  & 158   \\
  Street Scene \cite{Street}          & 46 & 35  & 205& 17  & 17    \\
  NWPU Campus \cite{cao2023new}          & 305& 242 & 43 & 28  & 17    \\
  \textit{UCF-Crime}$^\ddagger$ \cite{MIR}    & 1,610  & 290 &  -  & 13  & 950   \\ \bottomrule
  \end{tabular}
  }
  \begin{tablenotes}
    \footnotesize 
    \item All datasets are available at the anonymous GitHub repository.
    \item $^\dagger$The frame rate is set to 15 fps. $^\ddagger$UCF-Crime is weakly-supervised..
  \end{tablenotes}   
\end{threeparttable}
  \vspace{-10pt}
\end{table}

The generalized datasets contain RGB video sequences captured from real-world scenes and are commonly used for traditional deep VAD model training and testing. The details are presented in Table~\ref{t-data}. In P2VAD models, these datasets are often encoded or transformed, with morphological abstraction, optical flow extraction, or skeleton detection applied. 

Unsupervised methods typically use only regular events from the training set to model spatio-temporal normality, with positive samples (anomalies) appearing only during testing. In contrast, weakly supervised datasets provide balanced positive and negative samples with binary labels (0 or 1) for each video, allowing for supervised anomaly detection at the frame level using video-level labels.

Subway~\cite{Subway}, a prominent benchmark for unsupervised VAD, includes two long videos recorded at subway entrances and exits, with anomalies such as gate-jumping and walking in the wrong direction. UMN~\cite{UMN} contains 11 videos from three different scenes (indoor and outdoor) and focuses on detecting group anomalies in human behavior, potentially overcoming occlusion issues that may hinder skeleton-based VAD models. The UCSD Pedestrian~\cite{ped} dataset consists of Ped1 and Ped2 subsets, captured from two different camera angles on a campus pathway. Ped1 presents challenges in modeling due to the varying size of the target as it moves along the path, while Ped2's perpendicular angle makes anomalies such as skateboarding or cycling on sidewalks more detectable.

CUHK Avenue~\cite{avenue}, also recorded on a university campus, consists of 16 training and 21 test videos. Privacy concerns are heightened in this dataset due to clear facial, texture, and color information. Most anomalies involve motion cues (e.g., wandering or throwing objects), which can be detected using optical flow or skeleton extraction. ShanghaiTech~\cite{FFP} is a large-scale, multi-scene VAD benchmark containing 317,398 frames across 13 scenes, with 158 anomalous events. The performance of unsupervised methods on ShanghaiTech is generally lower compared to CUHK Avenue~\cite{avenue} and UCSD Ped2~\cite{ped}, likely due to label-independent noise in complex scenes, where skeleton-based VAD may offer better adaptation.

Street Scene~\cite{Street} captures common behaviors in traffic scenarios, such as jogging and parking, and contains 205 anomalous events across 17 categories. It also includes variations in lighting and provides bounding box annotations for anomalies. NWPU Campus \cite{cao2023new} captures events associated with different categories in various scenarios, yet these events may share similar appearances.

UCF-Crime~\cite{MIR}, the first weakly supervised VAD benchmark, includes 1,900 real-world surveillance videos, with 1,610 used for training and 290 for testing. Video-level annotations indicate whether each video contains an anomaly, but the exact temporal location is unknown. Anomalous events are categorized into 13 types (e.g., shootings, robberies), though all anomalies are labeled with 1, without finer-grained labels to differentiate behaviors from typical activity recognition tasks. UCF-Crime's human-related videos have been skeleton-annotated to train P2VAD models~\cite{boekhoudt2021hr}. ShanghaiTech's weakly supervised dataset~\cite{GCLNC} reorganizes the original ShanghaiTech dataset to include 437 videos.

Generalized datasets often contain detailed, identity-recognizable texture information, which can raise privacy concerns and make the data sensitive to noise. Studies have demonstrated that such appearance-based data do not significantly affect human-centric anomaly detection, and skeleton-based VAD approaches often perform better in terms of scene adaptability. Many P2VAD models have been validated on generalized datasets that have been preprocessed to fit privacy requirements. System-level P2VAD approaches generally assume that local VAD models operate in secure environments, thus continuing to use generalized datasets for training and testing.

\begin{figure}[t]
  \centering
  \includegraphics[width=\linewidth]{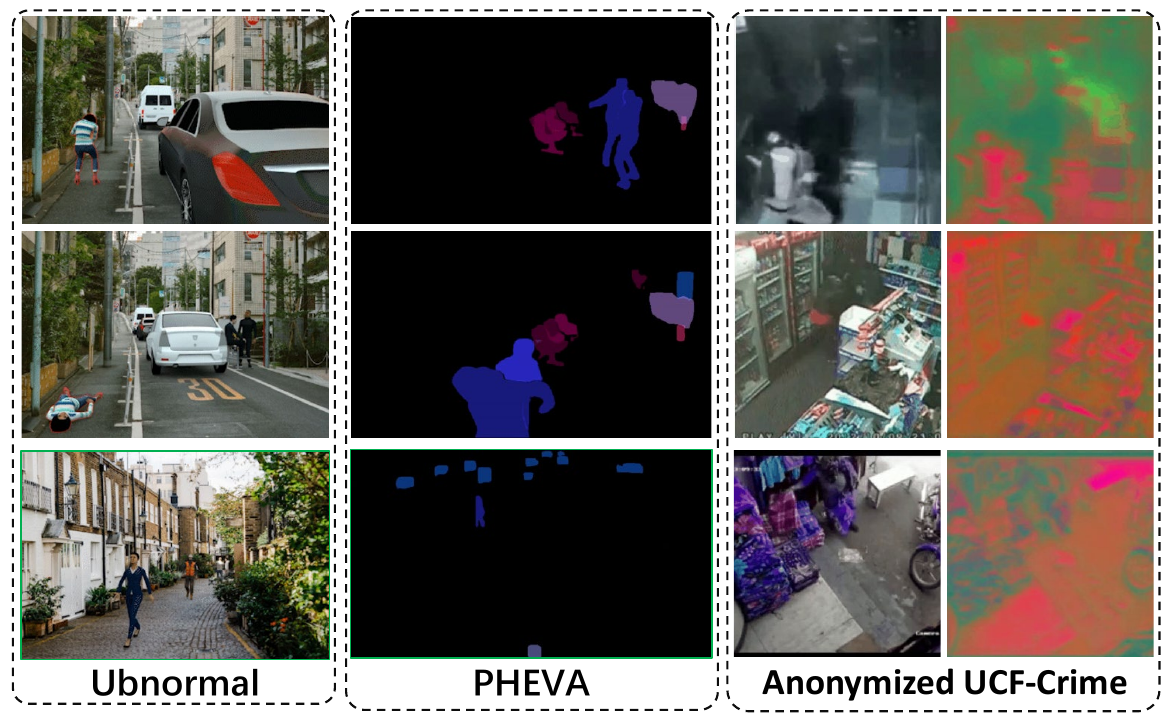}
  \caption{Examples of P2VAD-specialized datasets, including the synthesised Ubnormal \cite{acsintoae2022ubnormal}, PHEVA \cite{noghre2024pheva}, and Anonymized UCF-Crime \cite{fioresi2023ted}.}
  \label{f-data}
\vspace{-10pt}
\end{figure}

\subsubsection{P2VAD-Specialized Datasets}
P2VAD-specialized datasets \cite{rodrigues2020multi,noghre2024pheva} are mainly derived from preprocessing existing datasets to meet privacy-preservation requirements, such as HR-ShanghaiTech \cite{morais2019learning}, HR-Avenue \cite{morais2019learning}, and HR-Crime \cite{boekhoudt2021hr}. As the anonymized UCF-Crime shown in Fig.~\ref{f-data}, Fioresi et al. \cite{fioresi2023ted} use U-Net to remove the privacy-sensitive information in the UCF-Crime \cite{MIR}, XD-violence \cite{HL-Net}, and ShanghaiTech \cite{FFP} to achieve P2VAD. However, these specialized datasets are often used in individual studies without becoming widely benchmarked, which limits the development of P2VAD.

For codec-compressed datasets, coding standards such as H.264 and HEVC transform RGB video frames into human-unrecognizable binary formats. For example, Guo et al.~\cite{8766853} used H.264/AVC reference software (version JM-18.6) to encode Subway \cite{Subway}, UMN \cite{UMN}, UCSD Pedestrian \cite{ped}, and CUHK Avenue \cite{avenue} datasets, rendering their appearance information unavailable, with a variable bit rate mode applied. Li et al.~\cite{6890212} focused on anomaly detection in HEVC-compressed videos, compressing 16 traffic accident videos and 20 normal traffic videos from the MIT traffic dataset \cite{wang2008unsupervised} using an IPPP coding structure.

While synthetic datasets such as Ubnormal~\cite{acsintoae2022ubnormal} are primarily designed for traditional RGB-based VAD models, they do not contain real identifiable information. as shown in Fig.~\ref{f-data}. Using a data generation engine allows for the creation of diverse anomalies. Ubnormal can thus be used to train P2VAD models that avoid appearance bias. This dataset consists of 248 training videos, 64 validation videos, and 211 test videos, totaling 660 synthesis anomalies.

Most skeleton-based VAD datasets are created from generalized VAD datasets using pose estimation models. Hirschorn et al.~\cite{hirschorn2023normalizing} applied Crowdpose~\cite{li2019crowdpose} and YOLOX~\cite{ge2021yolox} to extract skeleton keypoints from the ShanghaiTech \cite{FFP} and Ubnormal \cite{acsintoae2022ubnormal} datasets. Additionally, Pose Flow~\cite{xiu2018pose} was used to track skeletons across video sequences. The PHEVA dataset~\cite{noghre2024pheva} uses YOLOv8 and HRNet~\cite{sun2019deep} for object detection and pose estimation, saving skeleton keypoints in the COCO17 format. The HR-Crime dataset~\cite{9959414}, extracted from UCF-Crime~\cite{MIR}, uses YOLOv3-spp~\cite{huang2020dc}, AlphaPose~\cite{fang2022alphapose}, and Pose Flow~\cite{xiu2018pose} for human detection, skeleton extraction, and skeleton tracking, respectively.

\subsection{Evaluation Metrics}~\label{sec62}

\begin{figure}[t]
  \centering
  \includegraphics[width=\linewidth]{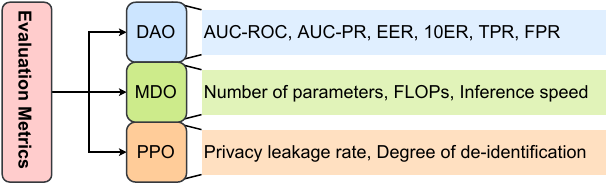}
  \caption{Evaluation metric system of P2VAD research, including detection accuracy-, model deployment-, and privacy preservation-oriented metrics.}
  \label{f-me}
\vspace{-10pt}
\end{figure}

The evaluation metrics for P2VAD can be classified into three categories: (a) Detection Accuracy-Oriented (DAO), (b) Model Deployment-Oriented (MDO), and (c) Privacy Preservation-Oriented (PPO), as shown in Fig.~\ref{f-me}. (a) and (b) are also commonly applied to evaluating RGB sequence-based VADs, which do not consider privacy concerns and focus on measuring a model's performance in distinguishing between normal and anomalous events. Since anomalous events are relatively rare in real-world scenarios, evaluation metrics that address imbalanced data, such as precision-recall curves, are typically used. Lightweight models and efficiency are critical factors determining a model's deployment feasibility, though direct comparisons are often difficult due to the lack of standardized experimental platforms. The primary concern of P2VAD is to enhance data security to mitigate public distrust of VADs, making it essential to assess privacy-preserving effectiveness quantitatively.

\subsubsection{Detection Accuracy-Oriented Metrics} 
The Receiver Operating Characteristic (ROC) curve illustrates the trade-off between the True Positive Rate (TPR) and the False Positive Rate (FPR) at different threshold settings and is commonly used to evaluate a VAD model's ability to distinguish between normal and anomalous events. TPR and FPR are as follows:
\begin{equation}
  \mathrm{TPR}=\frac{TP}{TP+FN}, \mathrm{FPR}=\frac{FP}{FP+TN},
\end{equation}
where $TP$ represents the number of correctly identified anomalous events, $FP$ the number of normal events misclassified as anomalous, $FN$ the number of missed anomalous events, and $TN$ the number of correctly identified normal events. The AUC-ROC (Area Under the ROC Curve) quantifies the overall discriminative capability of the model. A value close to 1 indicates strong performance in distinguishing between normal and abnormal events. However, AUC-ROC can be sensitive to imbalanced data, as it does not directly account for the False Negative Rate (FNR), or missed anomalies.

The Precision-Recall (PR) curve is often more informative than the ROC curve in imbalanced data scenarios. Recall ($R$) is equivalent to TPR, while precision ($P$) is defined as:
\begin{equation}
P = \frac{TP}{TP+FP}.
\end{equation}

AUC-PR, the area under the PR curve, is suitable for evaluating model performance when there is a significant disparity between positive and negative sample sizes. In VAD scenarios where anomalies are rare, AUC-PR offers a more reliable measure of a model's ability to detect these rare events.

The Equal Error Rate (EER) is the rate at which the false positive and false negative rates are equal, reflecting the balance between sensitivity and specificity at a particular threshold. EER is especially meaningful in video surveillance systems where a balance between detecting anomalies and recognizing normal events is necessary. EER is defined as:
\begin{equation}
EER = FPR = FNR.
\end{equation}

Another related metric is the 10\% Error Rate (10ER), which is used when a specific false alarm rate is acceptable in video anomaly detection. It measures the false positive rate when the miss detection rate is constrained to 10\%. This metric, derived from the FMR100 in biometrics, is particularly useful in applications where strict control over detection error rates is required. It is defined as:
\begin{equation}
FNR = 10\%, \quad FPR = 10ER.
\end{equation}
10ER is especially relevant in security monitoring systems with stringent requirements for anomaly detection.

\subsubsection{Model Deployment-Oriented Metrics}
The number of parameters, FLOating-point Operations (FLOPs), and inference speed are key indicators of a model’s suitability for deployment. Specifically:
\begin{itemize}
    \item Number of Parameters: This refers to the number of learnable parameters in the model. Fewer parameters generally indicate lower complexity, making the model more suitable for deployment on devices with limited computational resources.
    \item FLOPs: The number of floating-point operations reflects the computational resources required during model inference. High FLOP counts may limit the model's feasibility for edge device deployment.
    \item Inference Speed: This is the average number of video frames processed per second during the inference stage, which indicates the real-time processing capabilities of the model. However, this metric is highly dependent on the experimental environment and hardware, making cross-model comparisons challenging.
\end{itemize}

\subsubsection{Privacy-Preservation Oriented Metrics}
In privacy-preserving VAD systems, the ability to protect sensitive information is as important as detection performance. The relevant metrics include:

\begin{itemize}
    \item Privacy Leakage Rate: It quantifies the percentage of the model’s output that can be reverse-engineered to reveal identity or sensitive information. Ideally, this rate should approach zero.
    \item Degree of De-identification: This measures the effectiveness of techniques like visual desensitization and feature abstraction in removing identifiable information. This can be indirectly assessed through image quality metrics such as Structural Similarity Index Measure (SSIM) or Peak Signal-to-Noise Ratio (PSNR). A higher degree of de-identification corresponds to lower similarity between the processed and original content.
\end{itemize}

\section{Discussion}~\label{sec7}

  \begin{table*}[!htbp]
   \centering
   \caption{Comparison of P2VAD Pathways.}
   \label{t-com}
   \renewcommand{\arraystretch}{1.2}
   \setlength{\tabcolsep}{6pt}
   \resizebox{\textwidth}{!}{%
   \begin{tabular}{>{\raggedright\arraybackslash}p{1.8cm}
   >{\raggedright\arraybackslash}p{3cm}
   >{\raggedright\arraybackslash}p{5.3cm}
   >{\raggedright\arraybackslash}p{6cm}
   >{\raggedright\arraybackslash}p{5.5cm}}
   \toprule
   \textbf{Pathway}   & \textbf{Basic Concept}       & \textbf{Pros}& \textbf{Cons}         & \textbf{Future Prospects}   \\ \midrule
   
   P2VAD with Non-Identifiable Elements  & In the post-data acquisition step, encryption compression, appearance abstraction, or NVLCs are used to remove personally identifiable information from the RGB sequence. & 
   \begin{itemize}
     \item Eliminates privacy concerns and enhances transparency at data source. 
     \item Encryption compression and appearance abstraction can work on existing videos. 
     \item Encrypted video reduces redundant background and increases inference speed by over 50 times. 
     \item NVLCs can function in extreme conditions, such as darkness and high-speed environments. 
   \end{itemize}  
   & \begin{itemize}
     \item Encryption retains only high-frequency components, limiting detection performance of appearance-motion joint anomalies. 
     \item Performance of appearance abstraction depends on additional methods (e.g., object detection and instance segmentation). 
     \item Non-visible light videos require specialized sensing equipment, increasing hardware and data preparation costs. 
     \item The ideas of RGB-based VAD methods are often incompatible with such data. 
   \end{itemize} 
   & \begin{itemize}
     \item Efficient video compression can enable the real-time P2VAD system. 
     \item Lightweight object detection and zero-shot segmentation models (e.g., SAM) can handle complex scenes. 
     \item Sensing and representation technologies with event cameras and infrared will further P2VAD development. 
     \item AI technologies (e.g., adversarial samples) can enhance data security and the robustness of P2VAD models. 
   \end{itemize} \\ \midrule
   
   P2VAD with Desensitized Intermediate Modalities  & In the pre-step of model learning, use pre-trained models such as skeleton key point calculation and smooth extraction to obtain intermediate data modalities without sensitive information. & 
   \begin{itemize}
     \item Existing RGB VAD datasets can be used for training after simply pre-preprocessing. 
     \item Input modalities can resist pixel noise and ensure robustness. 
     \item Skeletons significantly reduce data size, accelerating P2VAD model training. 
     \item Optical flow removes background interference while capturing fine-grained changes. 
   \end{itemize}  
   & \begin{itemize}
     \item Relies on keypoint extraction and tracking models, increasing computational cost. 
     \item RGB sequences with identifiable information are still needed for collection, raising trust issues. 
     \item Skeletons are limited to human-related anomaly behavior detection. 
     \item Optical flow struggles with occlusion and has lower information density than skeletons and RGB sequences. 
   \end{itemize} 
   & \begin{itemize}
     \item Improved skeleton extraction and tracking models can boost the performance of skeleton-based P2VAD. 
     \item Motion vectors could replace optical flow for dynamics descriptions. 
     \item Fusing multiple desensitized modalities could help detect diverse anomalies. 
     \item Multi-stream modeling and causal learning in conventional VAD can transfer to P2VAD with DIM. 
   \end{itemize} \\ \midrule
   
   Edge-Cloud Intelligence Empowered P2VAD    & In VAD systems for real-world applications, privacy computing, edge-cloud collaboration, and federated learning are used to eliminate security risks in data transmission. & 
   \begin{itemize}
     \item Secures information transfer between clients and servers. 
     \item Suitable for large-scale VAD systems at city levels with thousands of devices. 
     \item Can utilize existing VAD models directly on local devices, focusing on privacy during data transmission. 
     \item Can tolerate various data modalities and qualities, achieving cross-scenario global anomaly detection. 
   \end{itemize}  
   & \begin{itemize}
     \item Heterogeneity and distribution of clients complicate model training. 
     \item Sensitive RGB data is still collected for modeling, requiring public explanation. 
     \item Encryption may increase communication and computational costs, making it challenging for mobile devices. 
     \item Extends beyond traditional VAD research, needing knowledge from computing and IoT communities. 
   \end{itemize} 
   & \begin{itemize}
     \item Advances in privacy computing and AI security could increase the credibility of P2VAD systems. 
     \item Customized federated learning can support robust, multi-client, scene-specific P2VAD. 
     \item Integration with MIE and DIM methods will enhance overall security. 
     \item Multimodal LLMs could improve transparency in P2VAD applications. 
   \end{itemize} \\ \bottomrule
   \end{tabular}
   }
\end{table*}

Sections~\ref{sec3}-\ref{sec5} have presented the implementation and development of various categories of P2VAD approaches that are motivated by different research objectives and underlying assumptions. While existing works have made notable progress in  data collection, model learning, and system deployment, significantly advancing privacy-preserving techniques in video anomaly detection, P2VAD, as an emerging interdisciplinary research topic, still faces unresolved challenges in real-world applications. Furthermore, the proliferation of intelligent video surveillance systems and the widespread popularity of short video applications (e.g., TikTok, Kuaishou, and Youtube) among Generation Z provide a broad market for P2VAD. Meanwhile, advancements in AI security and IoT technologies are expected to deeply integrate with VAD research, significantly driving the development of P2VAD. The brief comparison of the strengths, weaknesses, and future directions of various P2VAD Pathways are outlined in Table~\ref{t-com}. Moreover, from the perspective of privacy preservation and the research of trustworthy P2VAD systems for large-scale video IoT, this section discusses the challenges and opportunities of P2VAD, analyzing the limitations of existing approaches and trends for future exploration.

\subsection{Challenges and Limitations}

\subsubsection{Trade-off between Detection Performance and Privacy Preservation} Traditional VAD research primarily focuses on optimizing accuracy without adequately considering the privacy sensitivities of users or the potential of models on resource-constrained devices. These methods typically rely on the performance-driven metrics introduced in Sec.~\ref{sec62} as the sole indicators of model effectiveness and superiority. Although some methods can achieve high AUROC scores (above 90\%) on RGB-based datasets, privacy concerns and limited inference speed hinder their real-world deployment. In P2VAD, techniques such as appearance abstraction (e.g., NVLCs videos, pose estimation, optical flow extraction) and encrypted video coding (e.g., H.264/AVC, HEVC) are employed to remove identity-sensitive information. While these methods effectively alleviate privacy concerns, they often result in degraded detection performance due to the loss of critical visual features. For instance, removing facial or bodily details through pose estimation can desensitize identity, but may also eliminate important contextual information necessary for detecting certain abnormal behaviors. This creates a tension between safeguarding privacy and maintaining robust detection capabilities. Additionally, encrypted video coding techniques render the original video data unrecognizable to human observers, which not only complicates anomaly detection but also diminishes the interpretability of the results.

\subsubsection{Computational Overhead in Heterogeneous Systems}
Urban-scale VAD systems often rely on edge intelligence techniques (e.g., mobile computing, edge-cloud collaboration) to optimize the allocation of computational resources and improve overall performance. However, addressing privacy preservation in such large-scale heterogeneous systems—comprising numerous devices such as surveillance cameras, smartphones, local gateways, data centers, and servers—not only requires efficient privacy-preserving computing strategies but also faces significant increases in computational cost. Edge computing typically requires sensitive raw RGB sequences to be processed locally on resource-constrained yet trusted client devices. Additionally, intermediate features often need to be encrypted before transmission to servers or global models to prevent data misuse during transmission. This increases the computational burden on local devices and affects the system's real-time detection performance. Deploying privacy-compliant VAD systems in scenarios requiring fast response and large-scale application remains a bottleneck, necessitating the development of specialized privacy-preserving computing techniques for video data and efficient offloading algorithms for video IoT devices.

\subsubsection{Limitations of Federated Learning} Pivacy-preserving distributed learning methods like federated learning have been introduced to P2VAD due to their ability to train models without sharing local data \cite{al2024collaborative}. However, when processing high-definition video data, frequent synchronization between client devices and the central server for model updates incurs significant communication overhead, which is unacceptable for mobile devices. Moreover, the heterogeneity among different cameras, smartphones, and video streaming clients presents significant challenges for FL-driven P2VAD systems. These devices differ substantially in the quality of captured data, video processing capabilities, and accessible network conditions, leading to unequal contributions to the global model. Developing effective global model update strategies and robust handling of missing local information is crucial. Lastly, existing federated VAD systems are still vulnerable to privacy leakage risks. Through advanced adversarial techniques such as model inversion attacks, attackers may reconstruct sensitive video data from shared model updates, compromising system privacy. While differential privacy has been proposed to mitigate this risk, it often comes at the cost of model accuracy and efficiency.

\subsubsection{Common Challenges from Emerging AI Technologies}
Generative models \cite{liu2025survey} like large language models (e.g., Generative Pre-Training \cite{radford2018improving} and Contrastive Language-Image Pretraining \cite{radford2021learning}) have been introduced to enhance VAD's ability to understand and describe anomalies \cite{tian2024supervised,wu2024vadclip,zanella2024harnessing,liu2025anomaly}. However, their human-like interaction capabilities \cite{yang2024stephanie} have further exacerbated public concerns about privacy in VAD. For example, when capturing supplementary semantics in RGB video sequences to improve VAD performance, large models tend to fully describe all visual information, including facial details, clothing appearance, gender, and ethnicity, without considering privacy implications \cite{wu2024open}. While spatial appearance and temporal motion cues are essential for VAD tasks, the presence of sensitive identity information is typically irrelevant and exacerbates public distrust. Furthermore, Embodied Artificial Intelligence (EAI) \cite{xu2024survey} and Explainable Machine Learning (XML) introduce new scenarios and research perspectives for P2VAD. EAI combines robots and perception devices capable of real-time environmental sensing and autonomous responses to emergencies, yet these systems must process large amounts of sensor data (e.g., RGB cameras, infrared cameras, and event cameras), making privacy concerns prominent. XML improves the transparency of P2VAD systems by allowing users to understand the model's detection process and decision logic. However, it may expose sensitive information, raising questions about balancing explainability and privacy preservation.

\subsection{Trends and Opportunities}

\subsubsection{Applications of Enhanced Privacy-Preserving Technologies} Recent advancements in AI security and privacy computing, such as homomorphic encryption and secure multi-party computation, hold promise for improving privacy preservation in VAD systems. Specifically, homomorphic encryption, which allows computations to be performed on encrypted data without decryption, enables encrypted video streams to be directly analyzed for anomaly detection, significantly reducing privacy leakage risks while maintaining detection performance. Additionally, video encoding techniques that preserve high-frequency information related to motion cues can achieve exponential speedups without compromising detection performance. By integrating homomorphic encryption, trusted real-time P2VAD becomes feasible. The incorporation of differential privacy into VAD models also offers potential for mitigating privacy leakage risks in federated P2VAD systems \cite{alnajjar2024anomaly}. Differential privacy ensures that individual video frames or data points cannot be reverse-engineered from shared model updates by introducing controlled noise.

\subsubsection{Edge-Cloud Collaboration P2VAD Systems}
Edge-cloud collaboration leverages the cloud's powerful computational capabilities while performing local data processing at the edge, allowing different layers of computing units to cooperate for maximizing detection benefits and resource utilization. This approach can offer scalable VAD solutions \cite{liu2025edge}. Additionally, only localized devices have direct access to raw data, enhancing system transparency and trustworthiness. Specifically, privacy-preserving algorithms execute on edge devices, and only encrypted or abstracted information is transmitted to the cloud for further analysis and model updates, reducing the risk of exposing sensitive video data while handling more complex large-scale anomaly detection tasks. As adaptive learning systems, especially online and continual learning techniques, continue to evolve, P2VAD systems are expected to improve performance in dynamic environments. These systems can continually update models based on new data, allowing them to adapt to evolving privacy threats and anomalous patterns. 

\subsubsection{Standardization and Ethical Considerations}
As P2VAD continues to develop, the standardization of privacy and security protocols is essential for guiding research and driving commercial applications. Standardized privacy measures and guidelines help ensure that VAD systems meet minimum privacy requirements and allow for consistent evaluation of system performance. Furthermore, collaboration between academia, industry, and policymakers is critical in establishing a set of ethical guidelines, particularly in sensitive domains such as public surveillance and healthcare. Beyond technical standards, privacy-specific evaluation metrics must also be more comprehensive and closely aligned with real-world concerns. Current evaluation frameworks focus primarily on anomaly detection performance and average inference speed. However, future systems must also consider privacy-specific metrics such as identity obfuscation levels, data minimization, and resilience against privacy attacks. Developing these metrics will be crucial for building trust in P2VAD technologies and promoting their widespread adoption.

\section{Summary}~\label{sec8}

In this article, we focus on the data security and privacy preservation of the spatial-temporal anomaly detection in surveillance videos for the first time, providing a comprehensive taxonomy for Privacy-Preserving Video Anomaly Detection (P2VAD). By systematically examining the primary concerns, fundamental assumptions, and modeling processes of existing work, we shed light on the development trends and intrinsic connections within this domain. We first investigate the potential privacy and security issues that arise during the data collection, model learning, and system deployment phases of VAD, summarizing the developmental trajectory and key challenges faced by existing approaches. Additionally, we gather research resources, including datasets and technical literature, to facilitate further exploration in P2VAD, which have been publicly available for readers to build upon. Finally, we discuss the remaining challenges and analyse future opportunities in P2VAD research. We hope this survey will serve as a standard reference and contribute to the advancement of P2VAD technologies, thereby enhancing public trust of VAD applications and ensuring their deployment for societal benefit.

\bibliographystyle{IEEEtran}

\bibliography{refs-short.bib}

\end{document}